\documentclass[lettersize,journal]{IEEEtran}
\usepackage{amsmath,amsfonts}
\usepackage{algpseudocode}
\usepackage{algorithm}
\usepackage{array}
\usepackage{graphicx}
\usepackage[caption=false,font=normalsize,labelfont=sf,textfont=sf]{subfig}
\usepackage{textcomp}
\usepackage{stfloats}
\usepackage{url}
\usepackage{verbatim}
\usepackage{cite}
\usepackage{rotating}
\usepackage{booktabs}
\usepackage{amssymb}
\usepackage{marvosym} 
\usepackage{comment}
\usepackage{tabularx}
\usepackage[dvipsnames]{xcolor}
\usepackage{booktabs}
\usepackage{multirow}
\usepackage{hyperref}

\begin{document}

\title{Cross-Subject and Cross-Montage EEG Transfer Learning for Robust Gait Adaptation Detection}

%\author{XXX,~\IEEEmembership{XXX, }
\author{Nicole Lai-Tan, Xiao Gu, Marios G. Philiastides, Fani Deligianni
        % <-this % stops a space
\thanks{We'd like to acknowledge funding from UKRI Centre for Doctoral Training in Socially Intelligent Artificial Agents, (EP/S02266X/1), EPSRC (EP/W01212X/1) and Academy of Medical Sciences (NGR1/1678). }% <-this % stops a space
\thanks{Manuscript received XX XX XXXX; revised XX XX XXXX.}}

% The paper headers
\markboth{Journal of XXXX,~Vol.~X, No.~X, XX~XXXX}%
{Shell \MakeLowercase{\textit{et al.}}: A Sample Article Using IEEEtran.cls for IEEE Journals}

\maketitle
\begin{abstract}
Personalised music-based interventions offer a powerful means of supporting motor rehabilitation by dynamically tailoring auditory stimuli to provide external timekeeping cues, modulate affective states, and stabilise gait patterns. Brain-Computer Interfaces (BCIs) thus hold promise for detecting EEG activity associated with adaptative and non-adaptive gait patterns, enabling dynamic modifications of rhythmic structures and the personalisation of such interventions across individuals.
However, inter-subject variability in EEG signals, further compounded by movement-induced artefacts and motor planning differences, hinders the generalisability of BCIs and results in lengthy calibration processes. 
We propose Individual Tangent Space Alignment (ITSA), a novel pre-alignment strategy incorporating subject-specific recentering, distribution matching, and supervised rotational alignment to enhance cross-subject generalisation.
Our hybrid architecture fuses Regularised Common Spatial Patterns (RCSP) with Riemannian geometry in parallel and sequential configurations, improving class separability while maintaining the geometric structure of covariance matrices for robust statistical computation.
Using leave-one-subject-out cross-validation, `ITSA' demonstrates significant performance improvements across subjects and conditions. The parallel fusion approach shows the greatest enhancement over its sequential counterpart, with robust performance maintained across varying data conditions and electrode configurations. 
The accompany code scripts can be found: \url{https://github.com/NLai-CDT/ITSA.git}
\end{abstract}


\section{Introduction}

Brain-computer interfaces (BCI) are effective tools for motor rehabilitation and understanding musical stimulus effects on motor function \cite{Benoit_2014, Moumdijian2018, Devlin_2019, LaiTan_2023}. %\added{The development of robust EEG processing methods that can generalise across individuals and recording configurations while maintaining practical applicability in rehabilitation settings is critical.} 
In stroke rehabilitation, BCIs decode the user's intention from brain electrical activity to provide sensorimotor feedback and enable control of external devices or motor functions \cite{Mane_2020, Xu_2020_PreAlignment}. Assistive BCIs for motor rehabilitation bypasses damaged neuronal pathways to provide alternative control of the external devices while rehabilitative BCIs  exploit neuro-plasticity by promoting the recovery of damaged pathways and therefore restoring impaired motor functions \cite{Mane_2020}.  Electroencephalography signals are often used for the input of BCIs as they provide portable, non-invasive, low-cost solutions and have high temporal resolution \cite{Orban_2022}. 

In this study, we decode EEG segments time-locked to adaptive versus non-adaptive auditory cues during rhythmically-cued walking.
Navigating complex environments requires dynamic gait adaptation through footfall timing modulation \cite{Ravi_2021}.  
Rhythmic Auditory Stimulation (RAS) integrates auditory pacing into gait training to promote synchronisation and motor adaptation \cite{Wagner, Wu_Kong_2022}.
%\replaced{In this study, we refer to the EEG activity associated with active gait adjustment following a change in the rhythmic structure of the auditory stimuli as the ``gait adaptation mental state''. This state is represented by EEG segments corresponding to the initial adaptive cues following the tempo change.}
These adaptive states are underpinned by distinct electrocortical signatures, including central and parietal beta-band modulations \cite{Wagner2016,Jacobsen_2024}, sensorimotor alpha and beta power decreases linked to superior adaptation capacity, and posterior medial frontal cortex (pMFC) step-cue delay negativity reflecting temporal prediction errors \cite{Wagner_2019}. Because these neural signatures capture locomotor adjustments, detecting them via a BCI is essential as it could enable personalized music-based interventions that deliver targeted sensorimotor feedback, bridging motor recovery, emotional regulation, and sustained user engagement \cite{Chang_2019, Mane_2020, Deuel_2025,LaiTan_2023}.

A considerable challenge in EEG-based BCIs is cross-subject generalisation, where models trained on one population must perform on entirely unseen individuals \cite{Saha_2020, Xu_2020_PreAlignment, He_Wu_2020, Bleuze_2022, Gu_2024}. Variability results from anatomical differences, electrode position shifts between sessions and inter-subject variations in brain functions \cite{Bleuze_2022, Saha_2020}. Additionally, auditory gait rehabilitation involves subjects in motion, which introduces movement artifacts and muscle activity that compromise EEG signal quality. Yet the sensitivity of BCIs to these motion-induced disturbances remains largely unexplored. This necessitates time-consuming calibration processes for new user making real-world deployment impractical \cite{Saha_2020, Mayaud__2016}. Consequently, reducing inter-subject variability while retaining discriminative information from EEG signals is a key methodological challenge for developing generalisable BCI systems.

To compensate for these inter-subject variability and consequent covariate shifts between training and testing feature distribution, transfer learning (TF) techniques have been proposed \cite{Saha_2020}. TF seeks to leverage known information from previous sessions and past participants - the \textit{source} dataset, to improve upon unseen samples of subjects - \textit{target} dataset \cite{He_Wu_2020, Bleuze_2022}. Pre-alignment strategies (PS) are crucial for optimising TF from the source to target domain \cite{Xu_2020_PreAlignment}.
However, current pre-alignment strategies face several critical limitations. First, many approaches perform alignment in Euclidean space and ignore the inherent geometric and curvature structure of covariance matrices that naturally lie on Riemannian manifolds \cite{He_Wu_2020}. Second, existing methods often applies global alignment strategies that do not account for individual subject characteristics adequately \cite{Bleuze_2022}, potentially leading to over-regularisation or insufficient adaptation that could be addressed with the integration of individual adaptation \cite{Rodrigues_2019}. Third, they focus on reducing variability across datasets to account for the limited available training data \cite{Xu_2020_PreAlignment} and typically ignore cross-domain scenarios with different channel configurations, such as training on high-density EEG data and testing on a portable system with fewer channels. 

These limitations motivate a pre-alignment strategy that accounts for subject-specific variability, preserves discriminative information, and adapts to differences in EEG representations across domains. Furthermore, the integration of spatial filtering techniques with geometric alignment methods remains underexplored, despite their complementary nature. Common Spatial Patterns (CSP) excel at maximising class separability \cite{RCSP}, while Riemannian geometry-based methods preserve the statistical structure of covariance matrices \cite{Barachant}. As such, fusion of CSP and Riemannian approaches could enable greater interpretability via spatial projections that reveal how discriminative features relate to neurophysiological processes, while also preserving the underlying covariant structure captured by Riemannian geometry.

In this work, we address these challenges by proposing \textit{Individual Tangent Space Alignment (ITSA)}, a novel pre-alignment strategy specifically designed to mitigate inter-subject variability in leave-one-subject-out (LOSO) classification scenarios. Our approach incorporates three key innovations: (1) subject-specific recentering that individually aligns each subject's covariance matrices, (2) distribution matching through feature rescaling, and (3) supervised rotational alignment with robust calibration procedures.
Additionally, we build on previous works \cite{LaiTan_2024} to introduce a hybrid feature extraction architecture that fuses Regularised Common Spatial Patterns (RCSP) with Riemannian geometry techniques in both sequential and parallel configurations. In the sequential approach, spatial filtering is applied to the signals first, followed by a tangent space projection to extract features. In contrast, the parallel approach independently derives the (1) features from spatially filtered signals and (2) features from the tangent space projection, and then concatenates them. This fusion approach aims to maximise class separability by enhancing discriminative features while simultaneously preserving the geometric structure of covariance matrices for more accurate statistical computations. 
We explore cross-domain variability arising from differing EEG electrode configurations between training and testing datasets, training on high-density and testing on simulated low-density electrode montages. 

Our evaluation framework employs rigorous leave-one-subject-out cross-validation with statistical validation, ablation studies, and interpretability analyses. Results demonstrate superior performance with `ITSA' and parallel RCSP-Riemannian fusion.

Section \ref{RelatedWorks} presents the related work, Section \ref{Preliminaries} briefly introduces notation, Section \ref{Methods} describes the mathematical background and methodology for our proposed pre-alignment strategy and feature extraction methods, Section \ref{CrossMontagesetup} describes the cross-montage data setup and feature dimensionality alignment, Section \ref{Results} presents performance results and ablation studies. Finally, we conclude with our discussions in Section \ref{Discussion}.

\section{Related Work}\label{RelatedWorks} 

\subsection{Therapeutic audio-cued and musical interventions in regulating affective states and 
 enhancing motor rehabilitation}

Music and audio-cued interventions effectively alleviate depressive symptoms and reduce stress by engaging the parasympathetic branch of the autonomic nervous system \cite{Leggieri2019, Axelsen2022}. 

Affective states are reflected in motor functions \cite{Deligianni2019}. For example gait features such as variations in step/stride lengths and gait speed can infer emotional states \cite{Deligianni2019, Adolph}, and serve as an observable indicator of mood disorders progression \cite{Feldman, Adolph, Kumar}. Motor coordination impairment is proposed to arise from dysfunctions in dopaminergic circuits, such as the basal ganglia-thalamo-cortical loops \cite{Feldman}.  
Affective states also influence gait initiation through priming action \cite{Naugle_2011}. The cognitive function that facilitates the postural adjustment period during the initiation of movement lies in the supplementary motor area (SMA), premotor cortex and basal ganglia, all of which have connections to the limbic structures that modulate affective states \cite{Naugle_2011}. Music-induced emotional contexts decrease anticipatory postural adjustments durations \cite{Doumbia_2023}, suggesting that modulation of affective states could lead to regulating gait initiation. 

Music-based motor rehabilitation exploits the link between motor and emotional regulation, as well as the relationship between motor and auditory processing networks \cite{ASAP}. 
The entrainment to external musical or auditory cues generates temporal expectations that regulate gait patterns \cite{Benoit_2014}. Continued musical interventions improve motor function by promoting neural plasticity by either strengthening affected neuronal motor connections or by triggering compensation mechanisms \cite{Domingos2021}. In particular, it is hypothesised that external sensory cues compensate for impaired time-keeping mechanisms by bypassing the affected basal ganglia circuits through the cerebello-thalamo-cortical networks\cite{Benoit_2014,Gonzalez_2019, Giorgi_2024}.

Models based on predictive coding \cite{Vuust} and oscillatory brain network interactions have been proposed to understand the top-down and bottom-up processes explaining motor-music entrainment \cite{Large2000, LargeP}. However these remain largely theoretical and challenging to observe in practice. On the other hand, BCIs present a way to record the cortical activities related to motor function improvements in response to auditory-cued stimulus. Publicly available datasets of brain recordings during different gait adaptation phases can provide insights into how humans optimise locomotion during rhythmic intervention. These insights can guide future investigations into \textit{how} and \textit{why} music-based interventions are effective, enabling more personalised and targeted rehabilitation strategies. Furthermore, music-based BCIs could facilitate home-based monitoring and earlier, more effective interventions, motivating the need for EEG-based BCI systems that can reliably characterise gait adaptation states while remaining robust to inter-subject variability and practical recording constraints.

\subsection{Geometric and Statistical Approaches to EEG Signal Processing}

Feature extraction and selection methods capture meaningful information and key characteristics, improving accuracy and robustness in EEG-based BCI systems \cite{Hameed_2025}.
Common spatial patterns (CSP), a well-established binary-class feature extraction method, improves separability between classes in motor imagery (MI) based BCIs \cite{RCSP}. CSP spatially filters raw signals by obtaining spatial features that maximise variance in one class while simultaneously minimising it in the other \cite{RCSP}. Among several CSP extensions, Filter bank CSP (FBCSP) is widely adopted \cite{Hameed_2025}.  FBCSP considers frequency band-specific spatial features and selects the most discriminative features across all bands \cite{Keng_2008}. While FBCSP improves performance by exploiting the concept that different EEG frequency bands would contain distinct task-related information, this may limit its generalisability across subjects. 
Despite reported success in improving classification performances, CSP relies on representative covariance matrix estimates ; therefore, it can be highly sensitive to noise. Furthermore, it has also been shown to overfit in small datasets \cite{RCSP, Reuderink_2008, Grosse_2009}. Therefore, Lotte and Guan \cite{RCSP} encourage the regularisation of either the CSP objective function or the spatial covariance matrix estimates. 

An alternative approach improves classification performance by directly exploiting the covariance structures, capturing statistical relationships between spatial regions \cite{Barachant_Multiclass}.  Covariance structures are by definition Symmetric Positive Definite (SPD) matrices, which form a differentiable manifold \cite{Barachant, Bleuze_2022}. Correct handling of these matrices relies on differential geometry, such as Riemannian geometry. Consequently, classification methods involving metric calculation (distance or mean) cannot be accurately approximated using conventional machine learning algorithms with Euclidean computations \cite{Barachant, Bleuze_2022}. Riemannian classifier such as Riemannian Minimum Distance to Mean (RMDM) implement feature classification directly on the Riemannian manifold while appropriately leveraging SPD covariance matrix curvature structures through Riemannian metrics \cite{Barachant_Multiclass, Bleuze_2022}.  Alternatively, conventional machine learning algorithms that rely on Euclidean metrics can be applied by projecting each covariance matrix from the manifold to a tangent space, enabling local Euclidean approximation \cite{Barachant}. 
Unlike interpretable CSP spatial features, Riemannian features pose a challenge to interpret physiologically \cite{Barachant_Multiclass}. Fusing CSP and Riemannian features can support interpretability while preserving the intrinsic geometric structures of covariance matrices. 

While these approaches provide complementary representations of discriminative and geometric EEG information, their performance can remain sensitive to inter-subject variability, motivating the use of transfer-learning and alignment strategies to improve generalisation across subjects and recording domains.

\subsection{Transfer learning in BCI} \label{PS_RelatedWorks}

High variability between individuals' EEG data creates substantial challenges for cross-subject and cross-dataset generalisation in BCI algorithms. Pre-trained models often exhibit significant performance degradation when applied to new subjects or even different sessions of the same subject \cite{Xu_2020_PreAlignment}. Transfer learning (TL) leverages data from other subjects or sessions to improve learning for a new subject or sessions, thereby reducing or eliminating the calibration session requirements \cite{Wu_2022, Zhuang_2021}. Long calibration sessions increase mental fatigue in users, reducing experimental effectiveness \cite{Huang_2021}. TL theoretical basis stems from the transfer generalisation theory, where knowledge or skills transfer between domains relies on generalising past experiences \cite{Zhuang_2021}.  Domains must be sufficiently connected as if the domains do not share enough similarities or, conversely, if they are too similar, TL might fail \cite{Zhuang_2021}.  High-domain similarity can occasionally become misleading, resulting in incorrect assumptions and interference that can hinder effective transfer learning \cite{Zhuang_2021}. 

Zanini \textit{et al.}~\cite{Zanini_2018} proposes a transfer learning approach through a `pre-alignment step', where each training covariance matrix is transformed by recentering the Riemannian centre to identity matrices, reducing covariate shifts. In other words, as a result of the recentering, the mean of the newly aligned covariance matrices will be equal to the identity matrix, therefore reducing individual differences between the subjects \cite{Zanini_2018, Xu_2020_PreAlignment, He_Wu_2020}. Alternatively, Xu \textit{et al.}~\cite{Xu_2020_PreAlignment} and He and Wu~\cite{He_Wu_2020} consider direct time-series transformation. He and Wu \cite{He_Wu_2020} align each individual subject prior to the aggregated training subjects' signals for classification. Their exploration of reference matrix choice showed positive effects using the Riemannian compared to Euclidean mean of the underlying experimental condition \cite{He_Wu_2020}.

Xu \textit{et al.}~\cite{Xu_2020_PreAlignment} propose an  `online pre-alignment strategy' (OPS) to align testing datasets prior to model's testing, eliminating the need for calibration and constant Riemannian mean re-calculation when introducing new subjects.  Whilst the direct alignment of EEG trials offers greater flexibility in classification models and lower computational cost, using Euclidean mean addresses only the covariate shift and it overlooks \textit{prior probability shift} (i.e., the distribution of class labels may vary between subjects) and \textit{concept shift} (i.e., the relationships between EEG features and labels may change across subjects). Consequently, even after alignment, per-class distribution may still have large discrepancies across subjects and less separability \cite{He_Wu_2020}.

Rodrigues \textit{et al.}~\cite{Rodrigues_2019} proposes overcoming covariate shift through Procrustes Analysis by matching the statistical distribution of covariance matrices using geometric transformations (translation, scaling and rotation). Beyond recentering, Riemannian Procrustes Analysis (RPA) adds stretching and rotation steps to match data spread and align class centres between source and target datasets \cite{Rodrigues_2019}. Bleuz\'{e}  \textit{et al.} \cite{Bleuze_2022} proposes an RPA-inspired PS operating directly on the tangent space (TS).  They introduce a closed-form solution for the rotation step using singular value decomposition in tangent space alignment, thus reducing computation time in comparison to its RPA counterpart \cite{Bleuze_2022}. Bleuz\'{e}  \textit{et al.} \cite{Bleuze_2022} also explores clustering instead of singular class-wise centres when obtaining the anchor points in the rotation step. Unlike He and Wu \cite{He_Wu_2020}, both Rodrigues \cite{Rodrigues_2019} and Bleuz\'{e}  \textit{et al.} \cite{Bleuze_2022} combines multiple training subjects without preserving individual differences. This approach could lose subject-specific structures as it treats individual signals as if they were obtained from a single subject, potentially causing covariance space distortion and misrepresentation through diluted class structures. 

\section{Preliminaries} \label{Preliminaries}

Common Spatial Patterns (CSP) provide spatial filters to transform the raw EEG signals, highlighting features that contribute more strongly to discriminate between mental states while suppressing less informative components. Spatial features, $w$ are obtained through joint diagonalisation and eigendecomposition of the class-wise covariance matrices, which yields eigenvalues and their corresponding eigenvectors or, spatial features ($w$). Mathematically, this is achieved by maximising the objective function, $J(w)$ in Equation \ref{ObjectiveFunction}, where $E_k$ and $C_k$ represent the raw EEG signals and class-wise covariance matrix for class $k \in \{1,2\}$, respectively. 

\begin{equation}
\label{ObjectiveFunction}
J(w)=\frac{w^T E_1^T E_1 w}{w^T E_2^T E_2 w} = \frac{w^T C_1 w}{w^T C_2 w}.
\end{equation}

Alternatively, we can exploit the statistical relationship encoded in the covariance matrices by leveraging their inherent geometric structure. This approach accounts for the inherent curvature structure of the covariance matrices through the projection and transformation onto the tangent space (TS). After this transformation, conventional machine-learning algorithms can be effectively employed in the tangent space, where Euclidean computations more accurately approximate the true geometric relationships \cite{Barachant}. This tangent space mapping assumes that data is symmetric positive definite which is true for covariance matrices \cite{Barachant}. In this work, the assumption is preserved as covariance matrices are used as SPD inputs to the model, rather than being directly predicted.

The projection of the covariance matrix is described by Equation \ref{ProjectionTangent}. 
\begin{equation}
    P_i = \log_{C_{\text{ref}}}(C_i)= C_{\text{ref}}^{1/2} \mathrm{logm}(C_{\text{ref}}^{-1/2} C_i C_{\text{ref}}^{-1/2}) C_{\text{ref}}^{1/2},
 \label{ProjectionTangent}
\end{equation}
where $P_i$ represents the projected matrix obtained through a logarithmic mapping corresponding covariant matrix, $C_i$, that is subsequently half-vectorised. The matrix logarithmic operator $logm(.)$ is defined as $logm(A) = VD'V^{-1}$, where $D'$ contains the logarithms of the diagonal entries of the eigenvalues  \cite{Barachant}. The definition of the matrix logarithm assumes that $A$ is symmetric positive definite, which is true for covariance matrices. The reference covariance matrix $C_{\text{ref}}$ is typically chosen as the geometric mean, also known as Fr\' {e}chet mean of all covariance matrices $C_i$ defined by Equation \ref{MeanSPD}. 

\begin{equation}
    C _{\text{ref}}=\mathfrak{G}(C_1, ..., C_I) = \arg\min_{\textbf{C}} \sum_{i=1}^I \delta^2_R (C_{\text{ref}}, C_i),
    \label{MeanSPD}
\end{equation}
where  $\delta^2_R$ denotes the use of Riemannian distance as defined in Equation \ref{RiemannDistance}
\begin{equation}
    \delta_R(C_1, C_2) = \|\mathrm{logm} (C_1^{-1} C_2) \|_\mathcal{F} = [\sum^N_{n=1} \log^{2} \lambda_n]^{1/2},
    \label{RiemannDistance}
\end{equation}
where $\lambda_n$ are the eigenvalues of $C_1^{-1}C_2$, and $|\cdot|_F$ denotes the Frobenius norm.
A more detailed technical discussion of both RCSP and Riemannian geometry-based feature extraction can be found in \cite{LaiTan_2024}.

\section{ Methodology} \label{Methods}

We segment EEG windows from adaptive and non-adaptive heel strikes to create corresponding trials, $E\in \mathbb{R}^{t \times e\times s}$ , where $t$ denotes the number of trials, $e$ the number of channels/electrodes and $s$ the number of time samples respectively. The spatial covariance matrix for the trial $i$ is calculated using $C_i = \frac{1}{s-1}E_i E_i^T$ where $E^T$ denotes the transpose of the EEG matrix for trial $i$. 

\subsection{Pre-Alignment Strategy for Cross-Subject Transfer Learning} \label{PS_MethodDescription}

\begin{figure*} 
\centering
\includegraphics[scale=0.28]{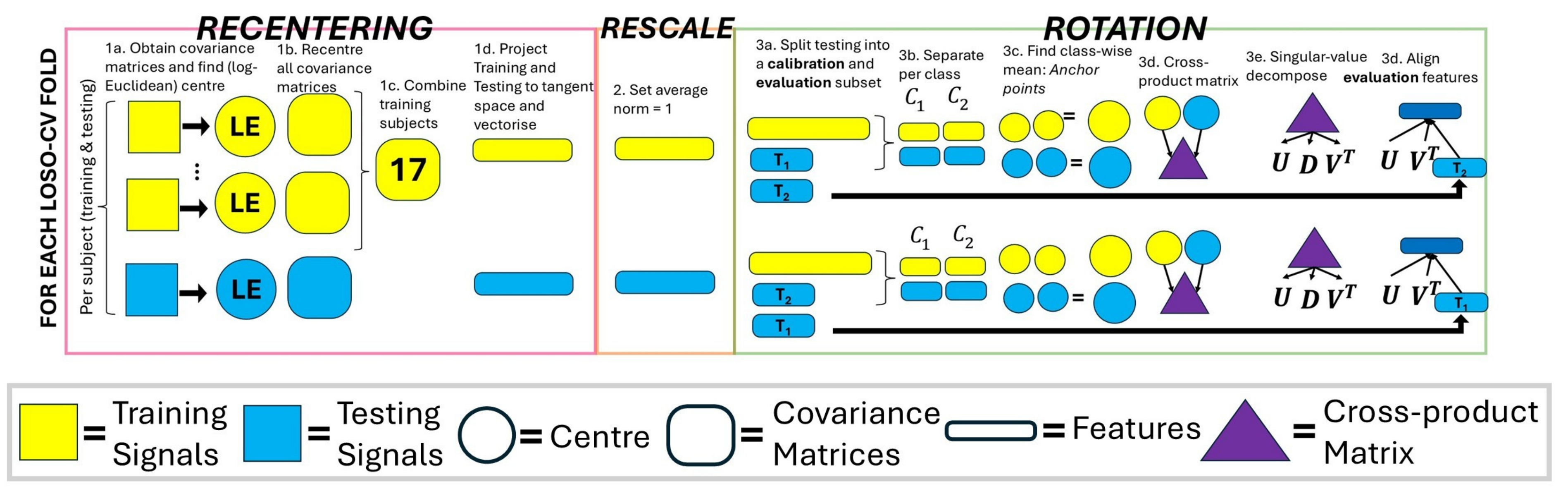}
\caption{Diagram illustrating the proposed PS - ITSA. \textbf{\textcolor{Goldenrod}{Training}} (yellow) and \textbf{\textcolor{blue}{Testing}} (blue) signals are represented as `squares', with corresponding covariance means as circles labelled with the metric used for calculation ( LE = Log-Euclidean).  Aligned signals are depicted as `squares with curved corners' and extracted features are displayed as `flattened' rectangles. The PS can be categorised into three steps (1) \textbf{\textcolor{magenta}{Recentering}} (pink), (2) \textbf{\textcolor{orange}{Rescale}} (orange) and (3) \textbf{\textcolor{ForestGreen}{Rotation} }(green). During the supervised Rotation step, only the calibration subset is used to derive transformational parameters to rotate the evaluation subset. A 2-fold cross validation is also implemented to ensure both subsets serve as the calibration and evaluation once.}
\label{PS_MethodImplementation}
\end{figure*}

We propose a novel pre-alignment strategy to optimise cross-subject transfer learning by independently aligning training and testing datasets. This approach aims to maximise the extraction of relevant features from training subjects that can be effectively transferred to testing subjects. 
A graphical representation of our proposed PS implementation - \textit{`Individual Tangent Space Alignment' (ITSA)} to improve cross-subject classification is demonstrated in Figure \ref{PS_MethodImplementation}. We extend the methods introduced by \cite{He_Wu_2020} and \cite{Bleuze_2022} by proposing a three-step alignment with an additional `\textit{calibration}' cross-validation. We first,  align each training subject individually to preserve subject-specific structures as in \cite{He_Wu_2020}.  Inspired by  Bleuz\'{e}  \textit{et al.} \cite{Bleuze_2022}, we include a further two alignment steps aimed at mitigating class-wise distribution shifts. We introduce an additional `calibration' cross-validation to prevent data leakage during the supervised `rotation' step.

\subsubsection{Subject-Specific Signal Recentering}
For each subject individually, we \textit{recentre} the covariance matrices to establish a common reference point across all subjects, following an approach similar to \cite{He_Wu_2020}. The subject-wise covariance matrix are recentred with respect to the global mean covariance matrix, $M$, such that the recentering transforms the global mean, $M$ to the identity matrix, $I_e$. Consequently, all subjects are represented relative to the same global reference point in the SPD space. Subsequent tangent space projection is implemented after the `recentred' covariance matrix,  $C_{\text{rec}}, $ as defined in Equation \ref{TS_projection_Identity}. Since the recentering is a congruence transformation of the SPD covariance matrix ($C$) \cite{Zanini_2018}, the resulting recentered covariance matrix $C_{\text{rec}} $, remains on the SPD manifold and can be subsequently mapped to the tangent space via the logarithmic map at the identity.

\begin{align}
    \hat{C}_{TS} = \text{Log}_{I_{e}}(C_{rec}) 
    &= I_{e}^{\frac{1}{2}} \log\left(I_{e}^{-\frac{1}{2}} M^{-\frac{1}{2}} C M^{-\frac{1}{2}} I_{e}^{-\frac{1}{2}}\right) I_{e}^{\frac{1}{2}} \notag \\
    &= \log\left(M^{-\frac{1}{2}} C M^{-\frac{1}{2}}\right)
\label{TS_projection_Identity}
\end{align}
where $M$ is calculated as the log-Euclidean mean of the subject's covariance matrices, $C$ represents the original covariance matrix, and $I_{e}$ is the identity matrix of dimension $e \times e$ (where $e$ is the number of channels). A half-vectorisation of the resulting tangent-space matrix then takes place to obtain a vector for subsequent processing.

\subsubsection{Distribution Matching via Feature Rescaling}
Inspired by  Bleuz\'{e}  \textit{et al} \cite{Bleuze_2022}, we include two additional alignment steps. First, we \textit{rescale} both the training and testing datasets to match their distribution around their respective centres of mass. This rescaling normalises the average norm within the recentred features space to one as demonstrated in Equation \ref{Rescaling}.  

\begin{equation}
    \tilde{C}_{SC} = \frac{\hat{C}_{TS}}{\frac{1}{t}\Sigma_n \parallel{\hat{C}_{TS_n}}\parallel}
    \label{Rescaling}
\end{equation}
where $t$ represents the number of trials and $|\cdot|$ denotes the Frobenius norm.

\subsubsection{Supervised Rotational Alignment with Calibration}
The second alignment step involves a two-stage process to\textit{ rotate} the testing vectors for optimal alignment with the training space. We first divide the rescaled testing features into two equal subsets ($T_1$ and $T_2$ in Fig. \ref{PS_MethodImplementation}), which represent a calibration subset and an evaluation subset, respectively. The calibration subset is used to estimate the class-wise calibration anchor points, while the evaluation subset is reserved for subsequent evaluation. These subsets corresponds to approximately 10 minutes of calibration for the `Advance Tempo' condition and 13 minutes for the `Delay Tempo' condition. This partitioning was conducted as a `block'-level to avoid data leakage between the calibration and evaluation subset as a result of the overlapping sliding windows.
Using the calibration subset, we derive the transformation parameters as follows. For a problem with $K$ classes, we calculate the mean of the rescaled vectors for each class $k$ in both the training set and the calibration subset, which we term \textit{anchor points}: 
\begin{equation}
\begin{aligned}
\bar{C}_{\text{train}_k} &= \frac{1}{N_{\text{train}_k}} \sum_{y_i = k} \tilde{C}_{{SC}\text{train}_i} \\
\bar{C}_{\text{calib}_k} &= \frac{1}{N_{\text{calib}_k}} \sum_{y_i = k} \tilde{C}_{{SC}\text{calib}_i}
\label{AnchorPoints}
\end{aligned}
\end{equation}
where $N_{\text{train}_k}$ and $N_{\text{calib}_k}$ are the number of trials for class $k$ in the training set and calibration subset, respectively, and $y_i$ represents the class label for trial $i$.

We then concatenate these class-wise anchor points to form training and calibration anchor matrices $\bar{C}_\text{train}, \bar{C}_\text{calib}$ :

\begin{equation}
\bar{C}_\text{train} = [\bar{C}_{\text{train}_1}, \ldots, \bar{C}_{\text{train}_K}], \quad
\bar{C}_\text{calib} = [\bar{C}_{\text{calib}_1}, \ldots, \bar{C}_{\text{calib}_K}]
\label{Concatenate_AnchorPoints}
\end{equation}

Next, we compute the cross-product matrix:
\begin{equation}
C_{TC} = \bar{C}_\text{train}\bar{C}_\text{calib}^T
\label{CrossProductMatrix}
\end{equation}

We decompose $C_{TC}$ using singular value decomposition:
\begin{equation}
C_{TC} = UDV^T
\label{SingularValueDecomposition}
\end{equation}

 The matrices $U$ and $V$ are truncated to retain the minimum number of features, $N_v$, that explain 99.9\% of variance obtaining $\tilde{U}$ and $\tilde{V}$. 
 This transformation is then applied to the evaluation subset, which was not used in determining the rotation parameters:

\begin{equation}
\hat{C}_{\text{ROT}_{\text{eval}}} = \tilde{U} \tilde{V}^T\tilde{C}_{\text{SC}_{\text{eval}}}
\label{AlignedTestingFeat_Multiplication}
\end{equation}

This two-stage approach ensures that the transformation parameters and the evaluation are derived from separate portions of the data, maintaining methodological rigour. 

Algorithm 1 in Appendix Section \ref{PsuedoCode_Section} provides a pseudocode for the proposed pre-alignment steps implemented on the training and testing datasets for each iteration of the leave-one-subject-out cross-validation.

\subsection{Fusion of Regularised Common Spatial Patterns and Riemannian Geometry for Enhanced Class Separability}
To further improve classification performances, we develop a fusion approach combining  Regularised Common Spatial Patterns (RCSP) with Riemannian geometry techniques, thus leveraging both spatial filtering capabilities and the inherent geometric structure of covariant matrices. 

\subsubsection{Regularisation of Covariance Matrices}
We regularise covariance matrices using Diagonal Loading to account for the smaller training dataset and avoid noisy contamination and overfitting \cite{RCSP}.  Regularisation  of the covariance matrix is performed by shrinking each covariance matrix towards the identity matrix shown in Equation \ref{Regualize_CovMatrix}, where $\tilde{C_k}$ is the regularised counterpart of the spatial covariance matrix, $C_k$ for class, $k$. $I$ is the identity matrix,  $s_k$ is a scalar scaling parameter. $\gamma \text{ and } \beta$ are the two user-defined regularisation parameters ($\gamma, \beta \in [0,1]$), and $G_k$ is the ``generic'' covariance matrix \cite{RCSP}. 

\begin{equation}
    \begin{split}
        & \Tilde{C}_k = (1- \gamma) \hat{C}_k + \gamma I \\
        \text{with} \text{ } &  {\hat{C}_k} = (1- \beta)s_kC_k+\beta G_k 
    \end{split}
    \label{Regualize_CovMatrix}
\end{equation}

Regularisation using the Diagonal Loading sets parameters $\beta= 0$ and effectively sets $G_k =0$. $\gamma$ is automatically obtained for each covariance matrix using the Ledoit and Wolf's method \cite{Ledoit_Wolf_2004}.

\subsubsection{Integration Approaches}
Fusing of RCSP with Riemannian geometry is explored in two approaches: `Sequential RCSP-Rie' and `Parallel RCSP-Rie'. 

\begin{figure}
    \centering
    \includegraphics[width=1\linewidth]{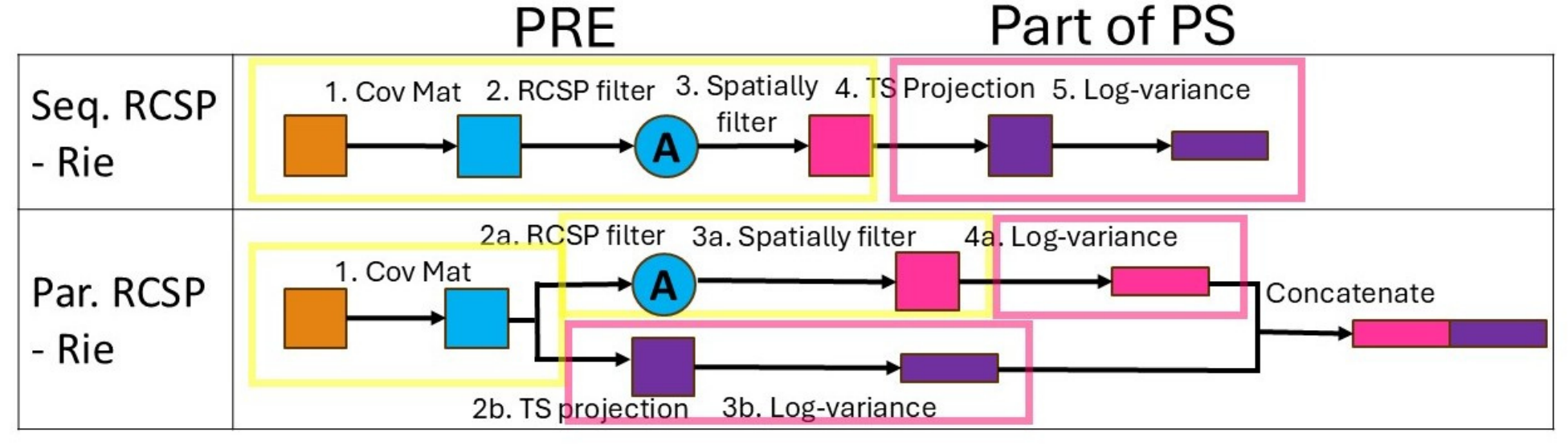}
    \caption{Diagram illustrating the fusion of RCSP - Riemannian geometry features. \textbf{\textcolor{Goldenrod}{Yellow outlines}} denote pre-PS steps and \textbf{\textcolor{magenta}{pink outlines}} represent steps that are part of the recentering step of `ITSA'. The keys are as follows: \textbf{\textcolor{orange}{Orange square}}  - EEG signals, \textcolor{blue}{\textbf{Blue square}}  - covariance matrices, \textbf{\textcolor{blue}{Blue circle}} - CSP filter obtained using arithmetic, depicted by the `A', \textbf{\textcolor{magenta}{ Pink square}}  - spatially filtered EEG signals, \textbf{\textcolor{Plum}{Purple square}}  - Projected dataset onto the tangent space, Rectangular  - extracted features in the form of taking log-variance. }
    \label{MethodImplementation_TS_PS}
\end{figure}

\paragraph{Sequential Integration (``Sequential (Seq.) RCSP-Riemannian Features'') }
`Seq. RCSP-Rie' combines the two techniques sequentially - firstly spatially filtering signals, followed by TS projection and (half) vectorisation to obtain input features. Since the proposed PS implementation includes a TS projection and subsequent vectorisation during the recentering step, spatial filtering is applied before the PS implementation, as illustrated by the yellow outline (steps 1 -3) in the top row of  Figure \ref{MethodImplementation_TS_PS} . Subsequent projection and vectorisation are then performed as part of the PS (recentering step), as shown by the pink outline (steps 4-5). 

\paragraph{Parallel Integration (``Parallel (Par.) RCSP-Riemannian Feature'') }
`Par. RCSP-Rie'  combines these two techniques in a parallel approach. Both spatially filtered features and tangent space features are obtained simultaneously and horizontally concatenated to obtain input features, as shown in the bottom row of Figure \ref{MethodImplementation_TS_PS}. Similar to in `Seq. RCSP-Rie', signals are spatially filtered before the PS implementation (steps 1-3a), and features are subsequently derived as part of the PS (step 4a). Tangent space features are obtained as a product of the implementation of the PS (steps 2b-3b) as part of the recentering step (pink outlines).  

\subsubsection{Classification with Support Vector Machines}

For the final classification stage, we employ support vector machines (SVMs) to discriminate between the two mental state classes. SVMs are particularly well-suited for this application due to their effectiveness with high-dimensional feature spaces and their robustness to overfitting when properly regularised.
The SVM classifier operates on the features generated from either the sequential or parallel integration approaches described in the previous section. This unified framework enables direct comparison between the different feature extraction methods while maintaining consistency in the classification stage. 
For all experiments, we use a linear kernel SVM with regularisation parameter set to the default parameter ($ C= 1.0$). 

\subsection{Leave-one-subject-out (LOSO) Calibration Process and Cross Validation}

We employed a leave-one-subject-out cross-validation (LOSO-CV) framework to evaluate cross-subject generalisation performance while preventing data leakage during the pre-alignment process. For each of the N subjects, one subject was held out completely for testing while the remaining N-1 subjects formed the training subjects. Initial subject-specific recentering was applied independently to each subject using only that subject's own data to calculate the Log-Euclidean mean. After recentering, training features from the N-1 subjects were concatenated to form the training set. Subsequent rescaled and rotated training features were obtained from this concatenated training set. Following the recentering and rescaling of the held-out subject's data, the features were divided into calibration and evaluation subsets, with rotation parameters estimated using a nested 2-fold cross-validation approach where each subset alternately served as calibration data to derive transformation parameters and evaluation data for performance assessment. Final performance was computed as the average across both folds, ensuring all testing data contributed to evaluation while maintaining separation between parameter estimation and performance evaluation.

\subsection{Dataset Description and Preprocessing}
We employed a publicly available dataset \cite{Wagner} (available at OpenNeuro.org with accession number ds00197144) to obtain our EEG signals for classification. The dataset contains EEG recordings of 108-channels from 20 healthy participants undergoing audio-cued walking tasks. The authors of the dataset \cite{Wagner} noted that two subjects (subjects 19 and 20) were heavily contaminated by artefacts, we subsequently removed them from further analysis. In addition, biomechanical data in the form of Electromyography (EMG) signals and kinematic recordings from 3 goniometers (hip, knee and ankle) are provided \cite{Wagner}. 

Participants would participate in a series of audio-cued walking tasks on a treadmill where they were asked to synchronise their heel strikes with rhythmic cues. After a period of steady-walking to a comfortable pace, the tempo of the auditory cue would suddenly increase or decrease - signifying the start of the `Advance' or `Delay' Tempo. Subsequently, participants would be required to adapt their gait pattern by lengthening or shortening their steps to synchronise to the new tempo. 

\begin{figure}
    \centering
    \includegraphics[width=0.95\linewidth]{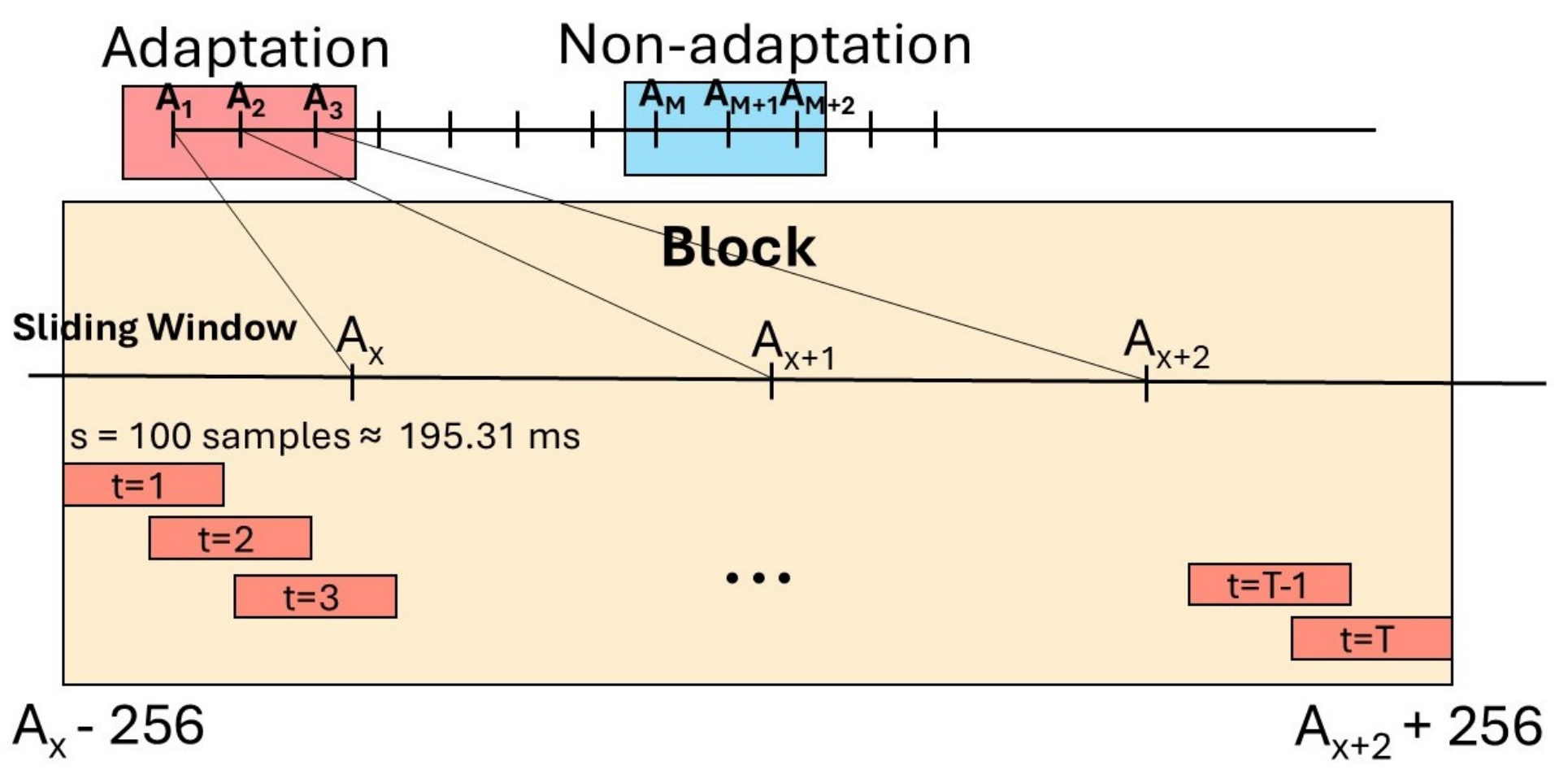}
    \caption{Sliding window $(s)$ segmentation of signals. The top horizontal line represents the EEG signal with vertical ticks marking the onset of auditory cues (AC) denoted as $A$. We segment adaptive and non-adaptive `blocks', defined as the first 3 adaptive (\textcolor{red}{red}) and middle (\textcolor{blue}{blue}) AC and illustrate as the orange rectangle taken from the [$A_x-256$: $A_{x+2} + 256$ ] of the adaptive condition. Windows of $s=100$ are slid across the segment for our dataset to obtain EEG trials, $t$}.
    \label{SlidingWindow}
\end{figure}

EEG signals are segmented into adaptive and non-adaptive auditory cues defined as blocks that encompasses the first and middle three auditory cues of each `Advance' and `Delay' tempo trials, $t$. This segmentation was motivated by the behavioural analysis reported in \cite{LaiTan_2024}, which showed that participants exhibited the greatest temporal deviation during the initial steps following the tempo change, with this response subsequently decreasing. We implemented a sliding window approach to optimise EEG's inherent non-stationary via overlapping, temporally localised analysis windows. We picked a window of $s =100$ samples (corresponding to 195.31 ms) with a 50\% overlap across the sectioned `block' of $[A_x-256:A_{x+2} + 256 ]$ around the three respective auditory cues, as visualised in Figure \ref{SlidingWindow}. As contralateral heel strikes occurred at intervals of approximately 530 and  750 ms for `Advance' and `Delay tempo' respectively, windows of $s=100$ offers a practical balance between temporal resolution and computational efficiency.

\subsection{Implementation Details}
We use \verb|pyRiemann| package \cite{pyriemann} for TS projection and subsequent half-vectorisation to handle the covariance matrix geometric structure. We leverage \verb|pyRiemann| integration with scikit-learn API for Riemannian preprocessing and SVM algorithms.

\section{Cross-Montage Setup} \label{CrossMontagesetup}

To simulate montage-specific testing datasets, we extracted channel-specific subsets from our original high-density 108-electrodes dataset, mimicking 10-10 and 10-20 montage configurations. Training datasets were not reduced to the target channel subset and training features were derived from the 108-electrode configurations. Corresponding RCSP filter and projected signals were derived using these testing subsets, while training datasets and respective RCSP filters and projected signals were obtained from the full 108-electrode configurations.
To ensure dimension compatibility and bridge dimensional differences between training and testing feature size prior to classification, feature reduction using PCA was applied to both training and testing features, independently. Positioning the feature reduction step in the final processing stages minimised information loss. Without `ITSA' implementation, feature reduction was applied to the fully-processed features post RCSP-Riemannian geometry integration (Seq. - purple rectangle and Par. - fused pink and purple rectangle in Figure \ref{MethodImplementation_TS_PS}).  However, with `ITSA', the supervised `rotation' step requires class-wise information from both training and testing features via the cross-product matrix. Therefore, feature reduction was applied following the `rescaling' step (within the orange outline of Figure \ref{PS_MethodImplementation}) to ensure dimensionality alignment.

\section{Results}\label{Results}

\subsection{Leave-one-subject-out (LOSO) classification}

The proposed pre-alignment strategy - \textit{`Individual Tangent Space Alignment' (ITSA)}, demonstrated substantial improvements in classification performance across most subjects for both feature extraction methods ( `Seq. RCSP-Rie' and `Par. RCSP-Rie') and temporal conditions (Advanced and Delay Tempos). 

\subsubsection{Performance Improvements by Condition}

In the advanced tempo condition, the proposed pre-alignment strategy enhanced performance for 10 subjects (subjects 1, 4, 6, 7, 10, 11, 12, 16, 17, and 18) in both the `Seq. RCSP-Rie' and `Par. RCSP-Rie' methods achieving an average F1 score 61.15\% and 61.34\%, respectively, representing a 56\% improvement rate across the cohort. 

In the delayed tempo condition,  the results showed differential improvements between methods. The `Seq. RCSP-Rie' improved performance for 8 subjects (subjects 3, 6, 12, 13, 14, 16, 17 and 18), while the `Par. RCSP-Rie' method demonstrated more robust improvements, benefiting 11 subjects (subjects 3, 6, 7, 10, 12, 13, 14, 15, 16, 17 and 18), achieving a 61\% rate. The averaged F1 performance for the `Seq.' and `Par. RCSP-Rie' approaches were 57.28\% and 58.52\%, respectively. 

Tables IV and VII in Appendix Section \ref{LOSO_IndSub} presents the individual F1 scores (\%) for each testing subject. 

\subsubsection{Statistical Analysis and Significance Testing}
Tables \ref{Baseline_WIndTSPS} present the LOSO-CV F1 scores (\%)  averaged across all subjects for the `Advance' and `Delay' conditions, respectively, comparing our proposed method `ITSA' PS against the baseline (where no pre-alignment strategy is implemented).

 Statistical analysis confirmed significant improvements across all tested configurations:
 
 \textbf{`Seq. RCSP-Rie':}  The Advance Tempo condition showed significant improvement ($t(17) = 2.3256, p = 0.03267$), while the Delay condition demonstrated even stronger significance ($W = 156.00, p = 0.0021$).

 \textbf{`Par. RCSP-Rie':} Both conditions yielded significant improvements, with the Advance condition achieving (Advance: $ t(17) = 2.6830, p = 0.01572$) and Delay condition ($t(17) = 4.0551, p = 0.0008$).

All statistically significant improvements($p<0.05$), using two-sided statistical tests, are indicated by `\textcolor{orange}{(*)}', corrected $p<0.0167$ using Bonferroni-correction are displayed by \textcolor{Goldenrod}{(**)} in Tables \ref{Baseline_WIndTSPS}. 

To determine statistical significance, we first performed normality testing using the Lilliefors test on the performance scores difference  \texttt{`ITSA' vs. Baseline}. 
The analysis showed that three of the four condition-method combinations followed normal distributions: `Seq. RCSP-Rie'  in the `Advance' conditions and `Par. RCSP-Rie' in both the `Advance' and `Delay' Tempo conditions.  For these normally distributed differences, paired t-tests were employed. However, the `Seq. RCSP-Rie' method in the Delay condition deviated significantly from normality (p = 0.0452), necessitating the use of the non-parametric Wilcoxon signed-rank test.

 \begin{table*}
        \centering
        \renewcommand{\arraystretch}{1.3} 
        \caption{Averaged F1-scores and standard deviation across testing subjects and standard deviations for classification of Advance and Delay Tempo features comparing the baseline (without PS) and our proposed `ITSA'. Metrics are reported for the two feature extraction approaches: Seq. RCSP-Rie and Par. RCSP-Rie.  95\% confidence intervals are also provided. Statistically significant improvements resulting from the PS  (uncorrected $p<0.05$ are  denoted by \textcolor{orange}{(*)}, corrected $p<0.0167$ using Bonferroni-correction are denoted by \textcolor{Goldenrod}{(**)} ).}
        \label{Baseline_WIndTSPS}
        \begin{tabularx}{\textwidth}{|l|X|X|X|X|} \hline
        & \multicolumn{2}{c|}{\textbf{Advance Tempo}} & \multicolumn{2}{c|}{\textbf{Delay Tempo}} \\\hline
        & Baseline (No PS) & LOSO (`ITSA' - \textit{proposed} PS) & Baseline (No PS) & LOSO (`ITSA' - \textit{proposed} PS) \\ \hline
        Seq. RCSP-Rie (\%) & 54.39 $\pm$ 11.05 (CI: 48.90 - 59.88) & 61.15 $\pm$ 7.27 (CI: 57.54 - 64.77) \textcolor{orange}{(*)} & 41.96 $\pm$ 18.10 (CI: 32.96 - 50.97) & 57.28 $\pm$ 5.65 (CI: 54.47 - 60.09) \textcolor{orange}{(*)} \textcolor{Goldenrod}{(**)} \\ \hline
        Par. RCSP-Rie (\%) & 56.23 $\pm$ 8.43 (CI: 52.03 - 60.42) & 61.34 $\pm$ 5.49 (CI: 58.60 - 64.07) \textcolor{orange}{(*)} \textcolor{Goldenrod}{(**)}& 42.65 $\pm$ 19.52 (CI: 32.94 - 52.36) & 58.52 $\pm$ 5.65 (CI: 55.71 - 61.33) \textcolor{orange}{(*)} \textcolor{Goldenrod}{(**)}\\ \hline
        \end{tabularx}
        \end{table*}

\subsection{Ablation Study of pre-alignment components}

To systematically evaluate the contribution of individual components within our proposed pre-alignment strategy, we conducted an ablation study comparing our  `Individual Tangent Space Alignment' (ITSA) method against two state-of-the-art pre-alignment baselines described in Section \ref{PS_RelatedWorks} that we refer to as: a) `Adaptive M' \cite{He_Wu_2020} and  b) `TS' \cite{Bleuze_2022}. The `Adaptive M' method proposed by He and Wu (2020) \cite{He_Wu_2020} aligns each subject's covariance matrices computed from the adapted signals through individual recentering as implemented in our proposed `ITSA' - however, unlike `ITSA', the alignment of signals is only performed through recentering and does not rescale or rotate features. To improve computational efficiency and enable the use of non-Riemannian machine learning algorithms, transformation and alignment of the covariance matrices occur in Euclidean space \cite{He_Wu_2020} using the arithmetic centre as a reference. Furthermore, higher performance was reported when computing the Euclidean mean exclusively over trials labelled as motor imagery, excluding rest trials. We adapted this strategy to align the training trials by taking Euclidean mean of trials from `adaptive' heel strikes - thus, the term `Adaptive M' is used. 

We also explored the effect that alignment in the tangent space has on the classification performance.  The `TS' method, proposed by  Bleuz\'{e}  \textit{et al}\cite{Bleuze_2022} initially recentres the training covariance matrices  choosing the log-Euclidean mean as the reference before projecting to the tangent space and subsequent half vectorisation. Two additional alignment steps, similar to `ITSA' - `rescaling' and `rotation' are implemented to ensure alignment of class means and consistency of dispersion around the mean. Table \ref{PS_AblationStudy}  provides a detailed breakdown of the components included in each method, facilitating a comprehensive comparison of their architectural differences. 

\subsubsection{Component-wise Performance Analysis}
\textbf{Advance Tempo Condition:} The analysis revealed that pre-alignment strategies failed to improve performance for only a limited number of subjects. Specifically, five subjects (subjects 3, 9, 13, 14 and 15) in the `Seq. RCSP-Rie'  method and two (subjects 3 and 13) in the `Par. RCSP-Rie' method did not show improvement with the use of any PS technique, representing failure rates of 28\% and 11\%, respectively. 
Tangent space alignment methods (both `TS' and `ITSA') achieved the highest performance, with optimal results for 10 out of 18 subjects (56\%) in both feature extraction methods (Averaged F1 scores: 57.44 (`Adaptive M') vs. 60.99 (`TS') vs. 61.15 (`ITSA') for `Seq. RCSP-Rie', and 59.29 (`Adaptive M') vs. 61.00 (`TS') vs. 61.34 (`ITSA') for `Par. RCSP-Rie').   

\textbf{Delay Tempo Condition:}
The `Delay Tempo' condition demonstrated more consistent improvements across pre-alignment strategies. Only subject 2 in the `Seq. RCSP-Rie' method failed to benefit from any pre-alignment approach, while all subjects in the `Par. RCSP-Rie' method showed performance gains with at least one strategy. 

Consistent with the `Advance Tempo' condition, tangent space methods achieved superior performance for the majority of subjects (8 out of 18 subjects (44\%) in 'Seq. RCSP-Rie' and 11 out of 18 subjects (61\%) in 'Par. RCSP-Rie'). Averaged F1 scores are reported as follows: 56.38 (`Adaptive M') vs. 57.13 (`TS') vs. 57.28 (`ITSA') for `Seq. RCSP-Rie', and 55.08 (`Adaptive M') vs. 58.45 (`TS') vs. 58.52 (`ITSA') for `Par. RCSP-Rie'.

Table \ref{Ablation_PS_results} presents the F1 scores across all subjects for the `Advance' and `Delay' conditions, respectively. All three pre-alignment strategies (`Adaptive M', `TS', and `ITSA') in the Delay tempo and the two tangent-space PS (`TS' and `ITSA') in the Advance tempo demonstrated statistically significant improvements over baselines (LOSO-CV without the use of PS) across both temporal conditions and feature extraction methods. 

Both `TS' and `ITSA' methods consistently achieved the highest classification performance, with statistical significant improvements compared to baselines across all conditions. 

The superior performance of tangent space methods compared to `Adaptive M' suggests that the primary benefit results from the additional rescaling and rotation steps in the tangent space rather than solely from individual recentering. 
Our proposed method, `ITSA', consistently outperformed `TS', demonstrating the added value of an initial subject-specific recentering around the mean before the tangent space projection, to allow generalisation in a LOSO setup.

\begin{table}
    \centering
    \caption{Ablation table on pre-alignment strategies. Checkmarks indicate the presence of the component in pre-alignment strategy.}
    \label{PS_AblationStudy}
    \begin{tabular}{|l|p{1.5cm}|p{1.5cm}|p{1cm}|p{1cm}|}  \hline 
         &  recentering (per subject) &  Rescaling Step & Rotation Step \\ \hline 
         Baseline (w/out PS) &  &  & \\ \hline 
         `Adaptive M' \cite{He_Wu_2020} & \makebox[1.5cm][c]{\checkmark} &  & \\ \hline 
         `TS' \cite{Bleuze_2022} &  & \makebox[1.5cm][c]{\checkmark} & \makebox[1cm][c]{\checkmark} \\ \hline 
         `ITSA' - \textit{proposed method} & \makebox[1.5cm][c]{\checkmark} & \makebox[1.5cm][c]{\checkmark} & \makebox[1cm][c]{\checkmark} \\ \hline
    \end{tabular}
\end{table}

 \begin{table*}
        \centering
        \caption{Averaged F1-scores and standard deviations across testing subjects for classification of Advance and Delay Tempo features from our LOSO-CV ablation study. We compared our proposed PS (`ITSA') with two PS baselines - `Adaptive M' \cite{He_Wu_2020} and `TS' \cite{Bleuze_2022}. Metrics are reported for the two feature extraction methods: Seq. RCSP-Rie and Par. RCSP-Rie. 95\% confidence intervals are also provided. Statistically significant improvement from the implementation of PS method  (uncorrected $p<0.05$ are  denoted by \textcolor{orange}{(*)}, corrected $p<0.0167$ using Bonferroni-correction are denoted by \textcolor{Goldenrod}{(**)} ).}
        \label{Ablation_PS_results}
        \begin{tabularx}{\textwidth}{|l|X|X|X|}\hline
    \multicolumn{4}{|c|}{\textbf{Advance Tempo}}\\\hline
             &LOSO (`Adaptive M') \cite{He_Wu_2020}& LOSO (`TS') \cite{Bleuze_2022}  &LOSO (`ITSA' - \textit{proposed} PS)\\ \hline 
             Seq. RCSP-Rie (\%)&57.44 $\pm$ 5.35 (CI: 54.78 - 60.10)& 60.99 $\pm$ 6.41 (CI: 57.80 - 64.18) \textcolor{orange}{(*)}&61.15 $\pm$ 7.27 (CI: 57.54 - 64.77) \textcolor{orange}{(*)}\\ \hline 
             Par. RCSP-Rie (\%)&59.29 $\pm$ 4.13  (CI: 57.24 - 61.35)& 61.00 $\pm$ 5.63 (CI: 58.21 - 63.80) \textcolor{orange}{(*)}&61.34 $\pm$ 5.49 (CI: 58.60 - 64.07) \textcolor{orange}{(*)} \textcolor{Goldenrod}{(**)} \\ \hline
     \multicolumn{4}{|c|}{\textbf{Delay Tempo}}\\\hline
     & LOSO (`Adaptive M') \cite{He_Wu_2020}& LOSO (`TS') \cite{Bleuze_2022}  &LOSO (`ITSA' - \textit{proposed} PS) \\\hline
     Seq. RCSP-Rie (\%)& 56.38 $\pm$ 6.41 (CI: 59.50-63.19) \textcolor{orange}{(*)} \textcolor{Goldenrod}{(**)} & 57.13 $\pm$ 5.50 (CI: 54.40 - 59.87) \textcolor{orange}{(*)} \textcolor{Goldenrod}{(**)} &57.28 $\pm$ 5.65 (CI: 54.47 - 60.09) \textcolor{orange}{(*)} \textcolor{Goldenrod}{(**)} 
    \\\hline
     Par. RCSP-Rie (\%)& 55.08 $\pm$ 9.25 (CI: 50.48 - 59.66) \textcolor{orange}{(*)} \textcolor{Goldenrod}{(**)} & 58.45 $\pm$ 5.65 (CI:55.64 - 61.26) \textcolor{orange}{(*)} \textcolor{Goldenrod}{(**)}  &58.52 $\pm$ 5.65 (CI :55.71 - 61.33) \textcolor{orange}{(*)} \textcolor{Goldenrod}{(**)} \\\hline
        \end{tabularx}
    \end{table*}
       
\begin{figure}
    \centering
    \includegraphics[width=1\linewidth]{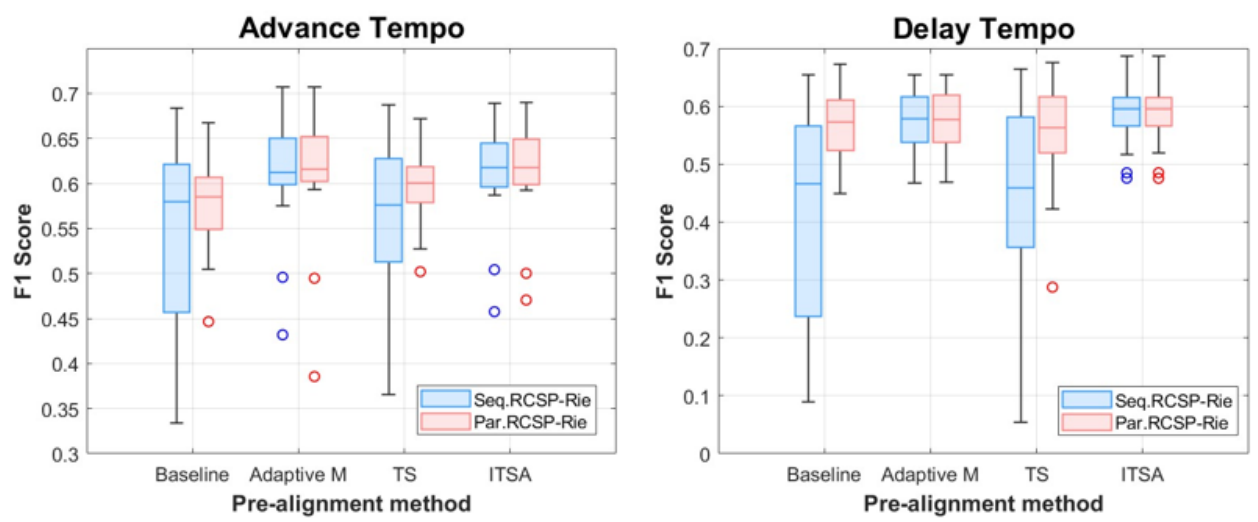}
    \caption{Box plot displaying the distribution of F1 score across 18 testing subjects and pre-alignment strategies during our LOSO-CV. Each box represents the interquartile range (25 and 75th percentiles) with the central line indicating the median. Whiskers extend to the most extreme points not considered to be outliers. Outliers are represented by `o' markers. We compared our proposed PS (`ITSA') with two PS baselines - `Adaptive M' \cite{He_Wu_2020} and `TS' \cite{Bleuze_2022}. Metrics are reported for the two feature extraction methods: \textcolor{blue}{Seq. RCSP-Rie} (blue) and \textcolor{red}{Par. RCSP-Rie} (red) and across the temporal conditions `Advance' and `Delay Tempo'.}
    \label{Boxplot_Ablation_PS_results}
\end{figure}

\subsection{Cross-Subject Interpretability}

\begin{figure*}[htbp]
    \centering
    \subfloat[`Seq. RCSP-Rie']{
        \includegraphics[width=0.45\linewidth]{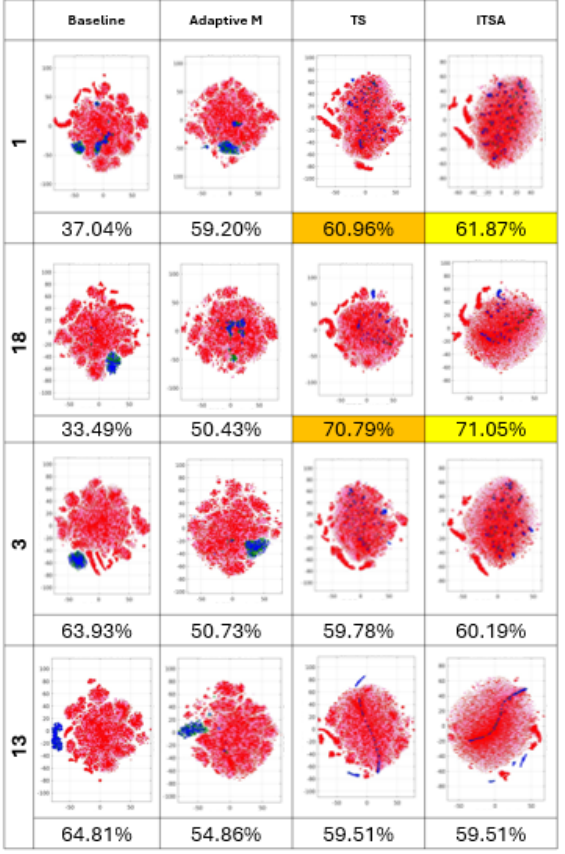}
        \label{TSNE_CompareSubjects_SpatFilt(A)}
    }\hfill
    \subfloat[`Par. RCSP-Rie']{
        \includegraphics[width=0.45\linewidth]{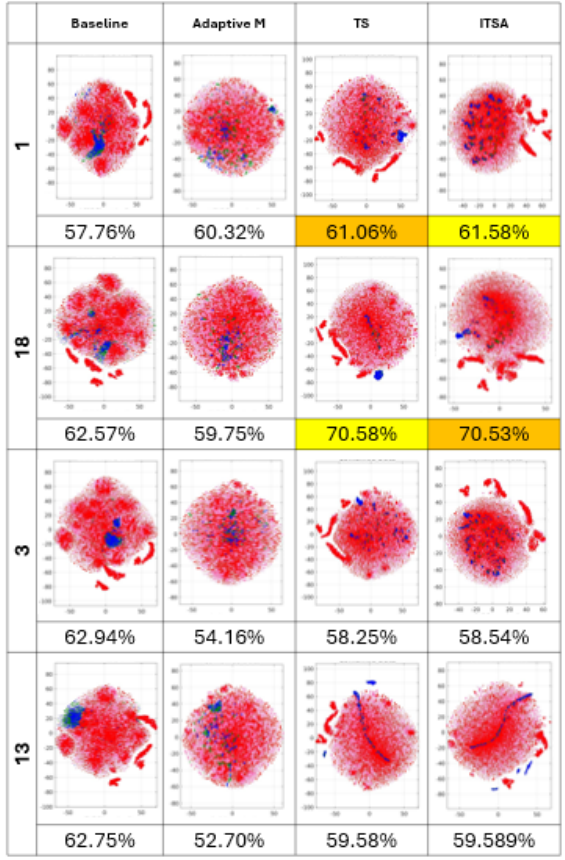}
        \label{TSNE_CompareSubjects_Concat}
    }
    \caption{Comparisons of testing subjects where `Adaptive M' , `TS' and `ITSA' PS enhanced or preserved classification performance with `Seq. RCSP-Rie' and `Par. RCSP-Rie'. The testing subject is denoted in the first column, with the first two rows categorised as `successful' (subjects 1 and 18) and the bottom two as `unsuccessful' (subjects 3 and 13). Respective F1 scores (\%) for each testing subject and applied PS are displayed beneath the t-SNE plot. The highest performing PS are highlighted in yellow, with the second  highest in orange. For `TS' and `ITSA' PS, corresponding t-SNE projections and F1 scores are reported using the `Fold 1' features of the PS calibration-testing used in the `Rotation' step. }
    \label{TSNE_CompareSubject}
\end{figure*}

We investigate whether subjects likely to perform poorly during testing, termed \textbf{`unsuccessful'}, can be identified in advance. In other words, these testing subjects can be interpreted as ones that do not generalise well to the population. `ITSA' reduces inter-subject variability by aligning each individual's features to a common centre and tangent space distribution. However, some subjects do not benefit from this, resulting in diminished classification performance, albeit with minimal deterioration.
Testing subjects were categorised as `unsuccessful' if no pre-alignment method (`Adaptive M' or `TS' or `ITSA' PS) improved upon baseline classification performance under either feature extraction approaches (`Seq.' or `Par. RCSP-Rie'). As `Delay tempo' yielded no `unsuccessful' subjects, subsequent analysis focuses on 'Advance tempo'. We compare two successful testing subjects (1 and 18) against two `unsuccessful' subjects (3 and 13). Feature distributions are visualised using t-SNE plots, comparing  `Adaptive M' \cite{He_Wu_2020} and `TS' \cite{Bleuze_2022} against the proposed `ITSA' PS. F1 scores (\%) for each PS and testing subject are shown below each plot for  `Seq. RCSP-Rie' (Figure \ref{TSNE_CompareSubjects_SpatFilt(A)}) and `Par. RCSP-Rie' (Figure \ref{TSNE_CompareSubjects_Concat}). For successful subjects 1 and 18, the highest and second-highest performing PS method is highlighted in yellow and orange, respectively. For `TS' and `ITSA' PS, t-SNE projections and F1 scores are from `Fold 1' of the calibration-testing split. Similar patterns were observed in `Fold 2'. 
t-SNE plots alone cannot reliably differentiate successful from unsuccessful PS outcomes as no consistent visual patterns align with PS performance across subjects.  In Figure \ref{TSNE_CompareSubjects_SpatFilt(A)}, `successful' subjects (1, 18) show greater dispersion and embedding of the testing features within the training feature space. However, `unsuccessful' subjects (3, 13) exhibit similar patterns, despite no performance gains. In Figure \ref{TSNE_CompareSubjects_Concat}, `Par. RCSP-Rie' projections appear more uniform, compared to their `Seq. RCSP-Rie' counterparts across all subjects. This uniformity may indicate  over-regularisation, potentially explaining `Adaptive M's' inferior performance compared to tangent space methods. Nonetheless, no clear visual distinctions separate 'successful' from 'unsuccessful' subjects, providing limited insight into factors driving PS effectiveness variation across subjects.

\begin{figure}
    \centering
    \includegraphics[width=0.85\linewidth]{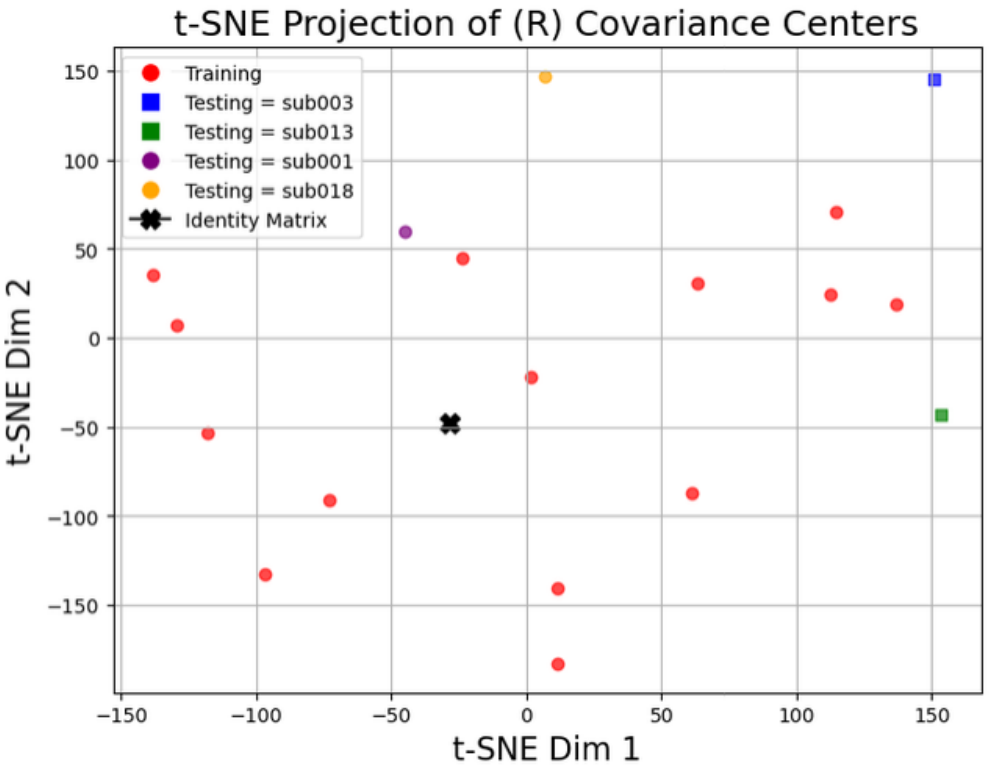}
    \caption{t-SNE projections (n components = 3, perplexity = 5) to visualise Riemannian distance between each subject's centre and the alignment reference point (identity matrix - \textbf{X}). Testing subjects of interest are displayed by blue (subject 3) and green (subject 13) squares, while successful testing subjects are displayed by purple (subject 1) and orange (subject 18) circles. The remaining subjects that create the training dataset are shown as red circles. Here, each sample corresponds to a subject-specific covariance representation, resulting in a total of $N=18$ samples.}
    \label{TSNE_RMean_Centre_TrainingTesting}
\end{figure}

Figure \ref{TSNE_RMean_Centre_TrainingTesting} shows t-SNE projections of the identity matrix (black cross), covariance centres for training subjects (red circles) and testing subjects 3 and 13 (blue and green squares, respectively). Both testing subjects appear at plot edges with the largest distances from the alignment reference point. This approach could help identify whether a subject's variability is suitable for effective PS implementation. 

\subsection{Cross-Montage Setup}

We simulated an experiment to determine if models trained on high-density EEG data (108-channels) could be applied to a lower-density configuration. The experiment is aimed to determine if pre-trained model can be used effectively in a portable EEG set up that would be more suitable for a real-world deployment, offering reduced complexity in set-up and computational demand. We chose channel subsets that corresponded to 10-20 and 10-10 montages (19 and 61 electrodes, excluding reference channels.  To determine the optimal number of features to retain without significant loss of performance - we conducted a feature reduction experiment as described in  Appendix Section \ref{FeatureReductionResults} (This analysis uses the 108-channel configuration and therefore does not provide direct cross-montage evidence). We deduced that a reduction of training features to retain 25\% and 1\% (for 10-10 and 10-20 montage) respectively would not cause significant loss in classification performances (Averaged performance loss - Seq: 3.47 (25\%) and 4.72 (1\%), Par: 3.14 (25\%) and 4.80 (1\%)). Feature reduction was implemented towards the end of the pipelines. Without the use of `ITSA', feature reduction was implemented on vectorised elements of spatially filtered covariant matrices that have been projected to the tangent space (i.e., processed features post RCSP-Riemannian geometry integration). This reduced the original dimensionality from 5886 and 5994 (for `Seq.' and `Par. RCSP-Rie' approach, respectively) to 1472 and 1498 features by retaining 25\% or  59 and 60 by retaining 1\%. As the rotation step in `ITSA' relies on class-wise information from both training and testing features via the cross-product matrix, feature reduction was recentred and rescaled half-vectorised tangent features (i.e., during the PS integration). This reduced the original dimensionality from 5886 and 11772 (for `Seq.' and `Par. RCSP-Rie' approach, respectively) to 1472 and 2944 features by retaining 25\% or  59 and 118 by retaining 1\%.

\begin{figure}[htbp]
    \centering
    \subfloat[]{
        \includegraphics[width=0.49\textwidth]{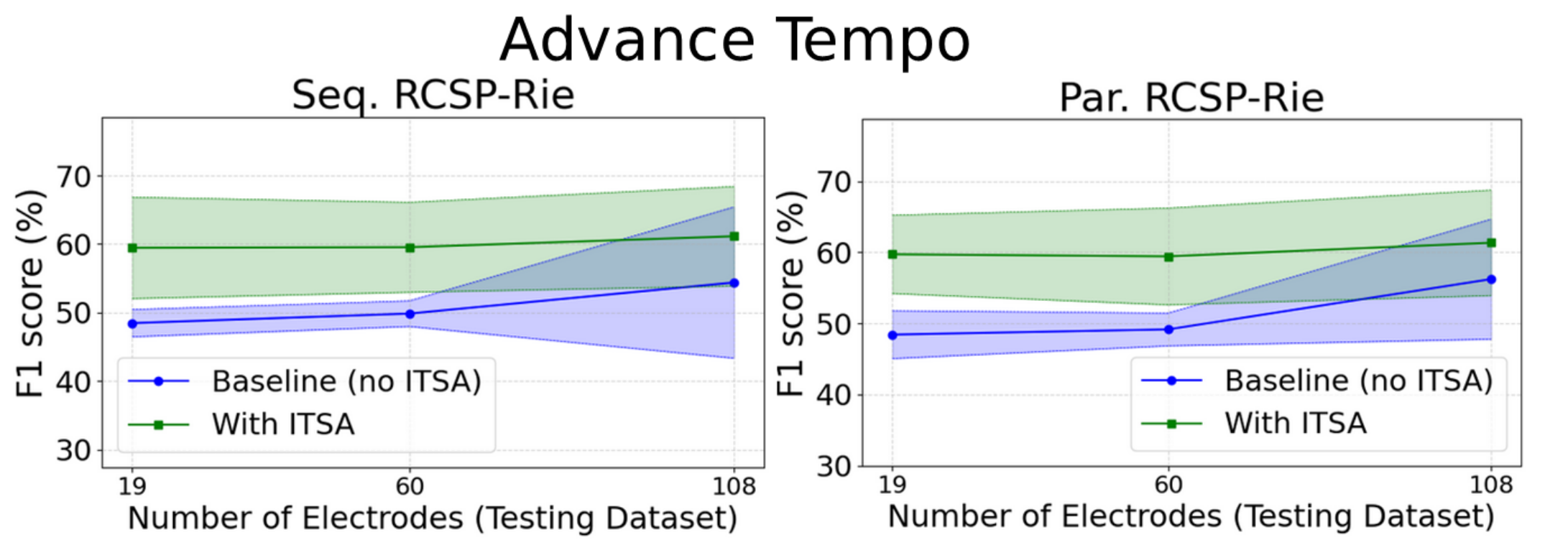}
    }
    \hfill
    \subfloat[]{
        \includegraphics[width=0.48\textwidth]{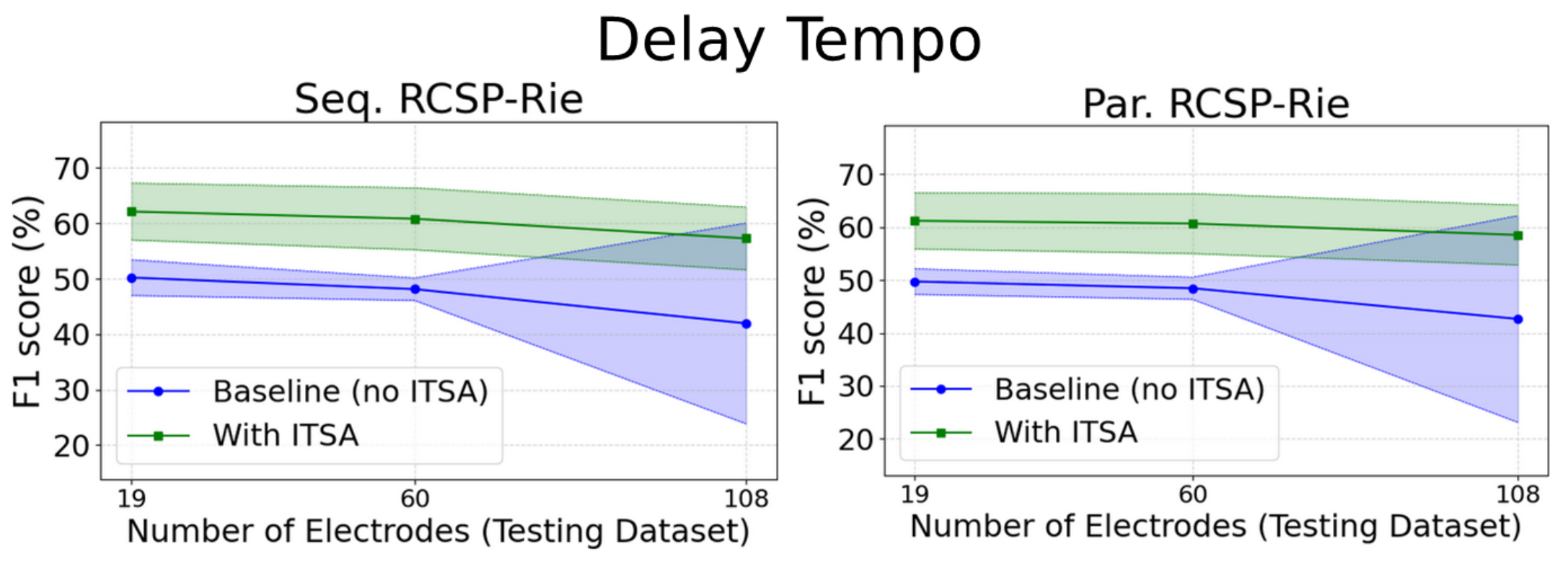}
    }
    \caption{LOSO classification performances for cross-montage experiment.
    Simulation of testing datasets using 61 and 19 electrodes (10-10 and 10-20 montage) was tested on the trained model with 108 electrodes. Comparisons between baseline performance (\textcolor{blue}{blue}) and after implementation of proposed `ITSA' (\textcolor{ForestGreen}{green}) are depicted. Shaded areas denote the corresponding standard deviations across leave-one-subject-out folds (i.e., across held-out test subjects).}
    \label{CrossMontageExperiment}
\end{figure}

Figure \ref{CrossMontageExperiment} depicts the LOSO-CV classification results for our cross-montage experiment. We simulate the training of our SVM model using high-density dataset of 108-electrodes, while testing on a lower-density 10-10 or 10-20 montage, consisting of 61 or 19-electrodes respectively. We employed PCA for feature reduction to ensure both training and testing dimensionality were consistent prior to input into the classifier. Training features were reduced to retain 25\% and 1\% of original features when simulating testing datasets from 10-10 and 10-20 montages. 

Our results demonstrate the potential of cross-montage experimentation, yielding promising outcomes. 
For both Seq. and Par. RCSP-Rie approaches and temporal conditions, applying `ITSA' across all three montages (108 electrodes, 10-10 and 10-20) consistently outperformed their respective baseline counterparts when `ITSA' was not implemented.
Overall, some loss of performance can be observed when reducing the testing dataset to lower montages in the Advance Tempo. Without the use of `ITSA', the average performance reduction was 4.54\% and 7.07\%  for the 10-10 montage  and  5.93\%  and 7.80\% for the 10-20 montage, when using  Seq. and Par. RCSP-Rie feature respectively. In contrast, implementing `ITSA' resulted in more consistent performance across both montages. Notably, average classification performances decreased by only 1.60\%  and 1.86\% for the 10-10 montage (61 electrodes),  and  1.69\% and 1.62\% 10-20 montage (19 electrodes) when using Seq. and Par. RCSP-Rie feature respectively.  

On the contrary, in the Delay condition, performance increased with the smaller testing montages, 10-20 montage (19 electrodes) and 10-10 montage (61 electrodes). Without the use of `ITSA', the average performance in the `Seq. RCSP-Rie' approach increased by 6.17\% and 5.8\% for the 10-10 montage and 10-20 montage, respectively. A similar observation is made in the `Par. RCSP-Rie' approach with an increase of 5.80\% and 7.07\% for 10-10 and 10-20 configurations. Similar to the Advance Tempo, `ITSA' implementation led to steadier performance across montages in comparison to baseline counterparts, with smaller increases of 3.55\% and 2.15\% for 10-10 montage and 4.85\% and 2.67\% increase in performance for 10-20 montage with respective Seq. and Par. RCSP-Rie features. 

Notably, `ITSA' pipeline maintained enhanced classification performance using only 19 or 61 electrodes, surpassing baseline performance achieved with a high-density test configuration for both Seq. and Par. RCSP-Rie approaches and temporal conditions. 
Note that the different PCA placement reflects the requirements of the two pipelines rather than an arbitrary preprocessing choice. ITSA requires both montages to share the same dimensionality after rescaling, so PCA is applied at that stage.

Our findings demonstrate the promising capability of leveraging high-density pre-trained models for effect classification of EEG signals from reduced-density montages such as the 10-10 and 10-20. Our results further support that the use of `ITSA' pipeline improves models' generalisability to unseen subjects even in lower-density configurations. Thereby, we believe the implementation of `ITSA' pipeline on a cross-montage experiment has the potential to reduce computational demand, set-up complexity, and in general improve practicality for real-world deployment. 

\subsection{Performance Curves}

To evaluate the scalability and robustness of our proposed pre-alignment strategy under varying data availability conditions, we analysed the relationship between training set size and classification performance. This analysis provides critical insights into the method's effectiveness in data-limited scenarios, which are common in clinical EEG applications.

\begin{figure}[htbp]
    \centering
    \subfloat[]{
        \includegraphics[width=0.48\textwidth]{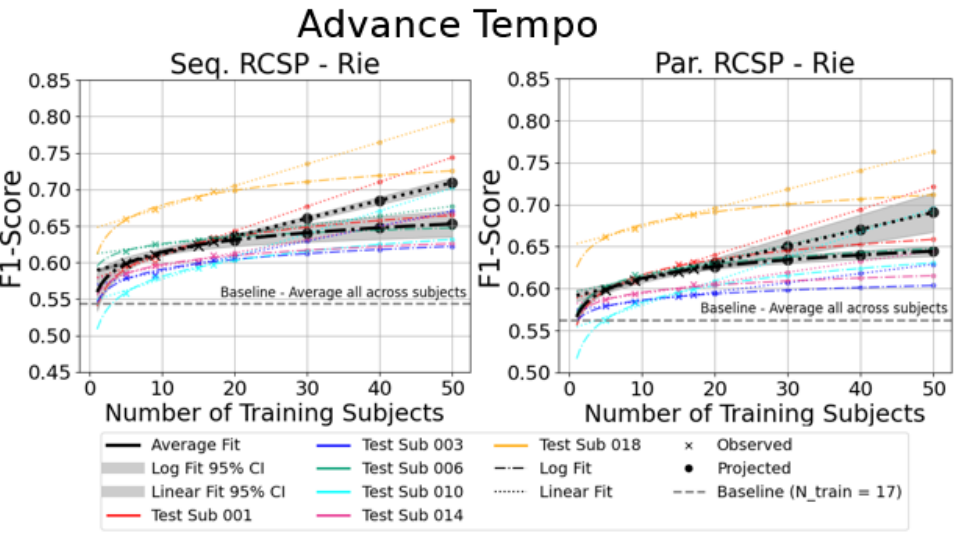}
    }
    \hfill
    \subfloat[]{
        \includegraphics[width=0.48\textwidth]{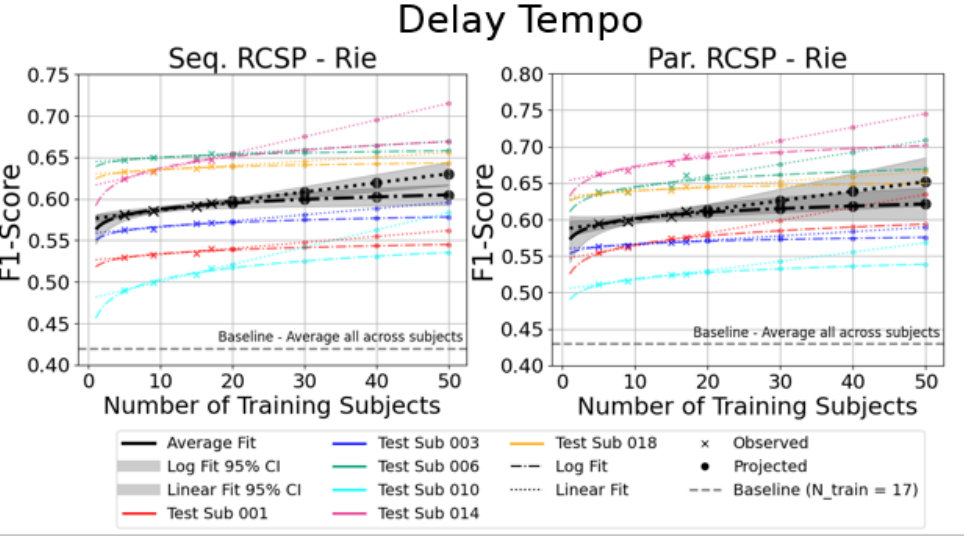}
    }
    \caption{Performance curves showing the relationship between the training size (number of subjects $N_\text{train}$) and averaged (black) LOSO-CV performance for test subject 1 (\textcolor{red}{red}), 3 (\textcolor{blue}{blue}), 6 (\textcolor{ForestGreen}{green}), 10 (\textcolor{cyan}{cyan}), 14 (\textcolor{magenta}{magenta}) and 18 (\textcolor{orange}{orange}) as testing subjects for (a) Advance and (b) Delay tempo. Observed F1 scores (crosses) are reported for $N_\text{train} = 5, 9, 15, 17$. Predicted F1 scores (solid dots) for a logarithmic (dash-dotted) and linear (dotted) fit are displayed. Baselines, without ITSA (grey dashed), are obtained as the averaged F1 score across all testing subjects, $N_\text{train} = 17$.}
    \label{LearningCurves}
\end{figure}

Figure \ref{LearningCurves} presents performance curves for both Advance and Delay tempo conditions, using subjects 1 (red), 3 (blue), 6 (green), 10 (cyan), 14 (magenta) and 18 (orange) as representative testing subjects in our LOSO-CV framework. The training size is represented by the number of training subjects ($N_\text{train}$), with the maximum available training size being $N_\text{train} = 17$ (utilising all remaining subjects after excluding the test subject).

For training sizes smaller than the maximum ($N_\text{train} < 17$), we implemented a robust 10-fold cross-validation procedure. In each fold, $N_\text{train}$ subjects were randomly sampled from the pool of 17 available training subjects. Our proposed pre-alignment strategy was applied consistently across all folds, as detailed in Section \ref{PS_MethodDescription}. Performance metrics for each $N_\text{train}$ value were obtained by averaging across the 10 folds to ensure statistical reliability. The baseline performance (grey dashed line) represents the averaged F1 scores across all LOSO-CV testing subjects without pre-alignment strategy implementation.

The performance curves display both observed and predicted performance metrics. Observed F1 scores are marked with crosses at discrete training sizes ($N_\text{train} = 5, 9, 15, 17$), while predicted metrics are represented by solid circles based on two projection models. The linear projection (dotted line) assumes a consistent linear relationship between training size and performance. The logarithmic projection (dash-dotted line) models the expected diminishing returns with increasing training size.

As anticipated, classification performance decreased consistently across all testing subjects as training size decreased. This trend was maintained for both temporal conditions, confirming the expected relationship between available training data and model performance.

Despite substantial reductions in training data ($>50$\% reduction, e.g., $N_\text{train} = 5$), the proposed method demonstrated remarkable resilience. Performance degradation remained minimal, with F1 score decreases of 5\% between all tested subjects examined. Furthermore, even at minimum training sizes, performance consistently exceeded baseline comparisons.

\subsection{Classification performance comparison across dataset splits}

We compare results across three dataset splits: `within-subject', `cross-subject' (as reported in \cite{LaiTan_2024}) and `LOSO'. As shown in Table \ref{DatasetSplit_Results}, performance decreases progressively across these splits for both Advance and Delay conditions.
            
For `within-subject' and `cross-subject' validations, training and testing splits were performed at the block level, ensuring all signals from a given block were assigned entirely to either set. This approach prevents overlapping or correlated signals from being shared across splits, mitigating the risk of data leakage.

The `within-subject' split exhibits the least inter-subject variability as training and testing data are obtained from the same subject. The `cross-subject' split pools data across all subjects, requiring the model to generalise more broadly. LOSO requires the greatest generalisation as the model is trained on a subset of subjects and tested on a completely unseen subject, yielding the poorest performance despite statistically significant improvements with `ITSA' PS for both conditions. This comparison highlights the challenge of inter-subject EEG variability.

\renewcommand{\arraystretch}{1.3} 

\begin{table*}[!ht]
\centering
\caption{F1-scores (\%) for classification of Tempo Advance and Tempo Delay features from within-subject, cross-subject, and LOSO splits. Metrics are reported for two fusion methods: Seq. RCSP-Rie and Par. RCSP-Rie.}
\label{DatasetSplit_Results}
\begin{tabularx}{\textwidth}{|l|X|X|X|X|} \hline
\multicolumn{5}{|c|}{\textbf{Advance Tempo}} \\ \hline
 & Within-subject & Cross-subject & LOSO (w/out PS) & LOSO (`ITSA' - \textit{proposed}) \\ \hline
Seq. RCSP-Rie (\%) & 94.18 & 68.17 & 54.39 & 61.15 \\ \hline
Par. RCSP-Rie (\%) & 87.95 & 69.64 & 56.23 & 61.34 \\ \hline
\multicolumn{5}{|c|}{\textbf{Delay Tempo}} \\ \hline
 & Within-subject & Cross-subject & LOSO (w/out PS) & LOSO (`ITSA' - \textit{proposed}) \\ \hline
Seq. RCSP-Rie (\%) & 94.33 & 69.96 & 41.96 & 57.28 \\ \hline
Par. RCSP-Rie (\%) & 88.73 & 70.73 & 42.65 & 58.52 \\ \hline
\end{tabularx}
\end{table*}

\section{Discussion and Conclusions} \label{Discussion}

In this study, we proposed a novel pre-alignment strategy - \textit{`ITSA'}, integrated with a fusion of Common Spatial Patterns and Riemannian geometry to improve the model's generalisability in a leave-one-subject-out cross-validation (LOSO-CV). Our approach combines subject-independent recentering to align all subjects to a common representational subspace with subsequent tangent space alignment steps that preserve the covariance matrix geometric structure.

The substantial performance degradation observed without pre-alignment highlights the negative impact of inter-subject variability on model generalisation. Our proposed method demonstrated consistent statistical improvements across all feature extraction methods and temporal conditions. To our knowledge, we are the first to demonstrate in a LOSO testing setup that through alignment and normalisation of EEG features across participants, inter-subject variability can be mitigated significantly. 

The ablation study demonstrated that subject-independent recentering and consequent rescaling and rotation steps in the tangent space, resulted in a more effective mitigation of inter-subject variability in comparison to current PS proposed in literature. 
We also observed some subject-specific variations in the effectiveness of the PS; however, we note that its application rarely resulted in significant performance deterioration in the `unsuccessful subjects. This suggests that our proposed PS is relatively robust and holds promise for generalisation across a diverse subject population. 

Crucially, our cross-montage experiments demonstrated the practical advantage of the ITSA-RCSP-Riemannian fusion framework. When training on high-density 108-electrode configurations and testing on reduced montages (10-10 and 10-20), the combined approach maintained superior classification performance compared to baseline methods. While our cross-montage experiment proposes a promising step towards montage-independent decoding strategies, it is important to highlight that real-world low-density headsets often employ dry electrodes with distinct noise and impedance challenges that is not considered in our simulated low-density configurations. In addition, the effectiveness of the proposed approach remains dependent on the available calibration duration, as reduced calibration time resulted in diminished performance gains in some conditions. This cross-montage generalisability addresses a critical limitation in BCI deployment, where training and testing setups may differ significantly in real-world applications.

Improving BCI systems generalisability, as demonstrated in our LOSO-CV and cross-montage deployment strategies with ITSA-RCSP-Riemannian integration, represents a critical step towards real-world BCI integration into music-based interventions. The ability to train on high-density electrode arrays while maintaining performance on lower-density montages significantly enhances practical deployment feasibility. It should be acknowledged that the current evidence is based on healthy participants in a controlled treadmill setting and simulated low-density montages. Further validation with clinical and consumer-grade EEG systems, realistic artefacts, and more diverse populations is therefore required to establish real-world generalisability. This approach in turn can then facilitate deeper investigations into the neural mechanisms that support auditory-cued motor responses. One method to studying cognitive processes underlying rhythmic entrainment involves disrupting steady-state behaviour and observing system's responses during return to stability \cite{Thaut_book_2013}. Neuroimaging evidence demonstrates that gait adaptation is associated with distinct spatiotemporal EEG patterns, including $\beta$-band activations in the frontal regions and power decreases in the central and parietal cortex are observed during gait adaptation \cite{Wagner2016} and the presence of step-cue delay negativity (SDN) that reflects the encoding of temporal prediction error \cite{Wagner_2019}. These findings support the interpretation of gait adaptation as a time-evolving cognitive-sensorimotor state, reflecting the dynamic integration of increased spatial attention, sensory integration and movement planning arising from locomotor adaptation \cite{Jacobsen_2024}. Generalisable BCIs can therefore support personalised musical interventions, where the musical stimuli are dynamically tailored to provide external timekeeping cues while modulating affective state, stabilising gait patterns and enhancing user's ability to initiate and adapt gait parameters. 

\bibliographystyle{IEEEtran}
\bibliography{Bib}

\section{Biography Section}
\vspace{-42pt}
\begin{IEEEbiographynophoto}{Nicole Lai-Tan} 
is currently a PhD student in the Social AI CDT at the University of Glasgow, funded by UKRI. She holds a MSc in Mathematics and a BSc in Mathematics and Music from the University of Leeds. Her research interest includes the development of AI algorithms to understand neurophysiological data with relation to gait dysfunction and its application within music therapy to support motor and cognitive rehabilitation. 
\end{IEEEbiographynophoto}
\vspace{-36pt}
\begin{IEEEbiographynophoto}{Dr. Xiao Gu} is a Senior Research Associate at the Computation Health Informatics (CHI) Lab, Department of Engineering Science at the University of Oxford. He holds a PhD degree in Computing and MRes degree (Distinction \& Dean's Prize) in Medical Robotics and Image Guided Intervention from Imperial College London in 2023 and 2019 respectively. His research focuses on integrating pervasive sensing and deep learning for healthcare, to address pressing generalisation issues in real-world applications. 
\end{IEEEbiographynophoto}
\vspace{-36pt}
\begin{IEEEbiographynophoto}{Professor. Marios G. Philiastides}
is a Professor at the School of  Psychology and Neuroscience and Director of the Centre for Cognitive Neuroimaging at the University of Glasgow.   He received his doctoral degree in Biomedical Engineering from Columbia University and MSc in Electrical Engineering from Stanford University. His research focuses on multimodal neuroimaging and computational modelling of human decision making and learning in health and disease. 
\end{IEEEbiographynophoto}
\vspace{-36pt}
\begin{IEEEbiographynophoto}{Dr. Fani Deligianni}
is a Senior Lecturer at the School of Computing Science at the University of Glasgow. She holds a PhD in Medical Image Computing (Imperial College London), an MSc in Advanced Computing (Imperial College London), an MSc in Neuroscience (University College London) and a MEng (equivalent) in Electrical and Computer Engineering (Aristotle University, Greece). Her research interest involves developing machine learning methods for the analysis of medical images, neurophysiological data and human motion while preserving the privacy or the users.  
\end{IEEEbiographynophoto}

\clearpage
\section{Appendix} \label{Appendix}

\subsection{Pseudocode of `ITSA'} \label{PsuedoCode_Section}

\begin{algorithm}
\caption{Pseudocode for proposed pre-alignment steps implemented on the training and testing datasets for each LOSO-CV fold}
\label{Alg:PSMethod}
\begin{algorithmic}[1]
\Require $E_\text{sub} \in \mathbb{R}^{t \times e\times s}$  \Comment{Windowed concatenated, standardised EEG signals for each subject $E_\text{sub}$ , for $t$ = number of trials,  $s$ = number of samples, $e$ = number of channels/electrodes}
    \State For  $ \forall\text{sub}, \forall t$, find spatial covariance matrix  
    \State Find Log-Euclidean centre of the set of training spatial covariance matrices. Centre covariance matrices and project to tangent space (TS) and vectorise using Equation \ref{TS_projection_Identity}
    \State Create testing and training features - holding one subject for testing and concatenate the remaining subjects to form the training features. 
    \State For  $\forall t$, rescale each both training and testing features using Equation \ref{Rescaling}.
\State \textbf{Training features:} Obtain Train\_AnchorPoints
\State \textbf{Testing features}: Partition rescaled testing data `\text{Test\_Rescale}' and corresponding label `\text{Y\_test\_dataset}' by block IDs into 50:50 subset 
\State Assign subset as either (1) calibration or (2) evaluation subset
\While {calibration subset} 
\State Using \textbf{calibration subset} derive \textit{Calibration} anchor points
\State Load \textit{Training} anchor points
\State Define cross-product matrix between the training and calibration anchor points (Equation \ref{CrossProductMatrix}) and decompose $C_{TC}$ using singular-value decomposition (Equation \ref{SingularValueDecomposition})
\State Perform Singular value decomposition  on cross-product matrix 
\State Find the $N_v$  vectors that account for 99.9\% variance, obtain $\tilde{U}$ and $\tilde{V}$. Align reserved testing vectors by matrix multiplication as shown in Equation \ref{AlignedTestingFeat_Multiplication} 
\EndWhile
\State \textbf{Evaluation subset}: Rotate left-out evaluation features by applying`\text{U\_tilde}' and `\text{VT\_tilde}' 
\State Test aligned testing features using model trained on  rescaled (and recentred) training features
\State Repeat to ensure both subsets serves as both `calibration' and `evaluation' once
\end{algorithmic}
\end{algorithm}
\subsection{LOSO-CV classification - individual subject } \label{LOSO_IndSub}

Tables \ref{Ind_LOSOCV_Advance} and \ref{Ind_LOSOCV_Delay} present the F1 scores (\%) for each testing subject in our LOSO-CV for both temporal conditions - `Advance' and `Delay Tempo'. 
\begin{table*}[t]
\centering
\caption{F1 scores (\%) for each testing subject in LOSO-CV classification - Advance Tempo. Highest performing score for each subject is highlighted in green.}
\label{Ind_LOSOCV_Advance}
\begin{minipage}{0.48\linewidth}
    \centering
    \renewcommand{\arraystretch}{1.2}
    \begin{tabularx}{\linewidth}{|p{0.6cm}|X|X|X|X|}
        \hline
        Test Sub & Baseline - w/out PS&LOSO (`Adaptive M') \cite{He_Wu_2020}& LOSO (`TS') \cite{Bleuze_2022}  &LOSO (`ITSA' - \textit{proposed} PS)\\\hline
        
        1& 37.04 & 59.20 & 62.43 & \textcolor{Green}{63.32} \\ \hline
        2& 62.16 & \textcolor{Green}{66.76} & 65.53 & 65.92 \\ \hline
        3& \textcolor{Green}{63.93} & 50.73 & 59.70 & 60.35 \\ \hline
        4& 52.77 & 56.86 & 60.06 & \textcolor{Green}{60.66} \\ \hline
        5& 45.72 & \textcolor{Green}{64.04} & 59.86 & 60.20 \\ \hline
        6& 57.36 & 60.01 & 62.90 & \textcolor{Green}{63.24} \\ \hline
        7& 64.19 & 62.05 & 65.33 & \textcolor{Green}{65.36} \\ \hline
        8& 44.45 & \textcolor{Green}{61.15} & 43.19 & 38.55 \\ \hline
        9& \textcolor{Green}{60.54} & 55.92 & 49.57 & 49.47 \\ \hline
        10& 33.37 & 44.65 & 57.55 & \textcolor{Green}{59.72} \\ \hline
        11& 68.37 & 60.72 & 70.70 & \textcolor{Green}{70.71} \\ \hline
        12& 52.57 & 59.74 & \textcolor{Green}{64.98} & 64.88 \\ \hline
        13& \textcolor{Green}{64.61} & 54.86 & 62.01 & 62.01 \\ \hline
        14& \textcolor{Green}{61.09} & 60.32 & 60.50 & 60.99 \\ \hline
        15& \textcolor{Green}{61.45} & 57.77 & 59.94 & 61.14 \\ \hline
        16& 57.74 & 55.25 & 64.66 & \textcolor{Green}{65.18} \\ \hline
        17& 58.17 & 53.55 & \textcolor{Green}{59.98} & 59.33 \\ \hline
        18& 33.49 & 50.43 & 68.95 &  \textcolor{Green}{69.70} \\ \hline
        \hline
        Avg& 54.39 & 57.44 & 60.99 & \textcolor{Green}{61.15} \\ \hline
        STD& 11.05 & 5.35 & 6.41 & 7.27 \\ \hline
    \end{tabularx}
    \caption{`Seq. RCSP-Rie'}
\end{minipage}
\hfill
\begin{minipage}{0.48\linewidth}
    \centering
    \renewcommand{\arraystretch}{1.2}
    \begin{tabularx}{\linewidth}{|p{0.6cm}|X|X|X|X|}
        \hline
        Test Sub & Baseline - w/out PS&LOSO (`Adaptive M') \cite{He_Wu_2020}& LOSO (`TS') \cite{Bleuze_2022}  &LOSO (`ITSA' - \textit{proposed} PS)\\\hline
        1& 57.76 & 60.32 & 62.84 & \textcolor{Green}{63.09} \\ \hline
        2& 65.55 & \textcolor{Green}{67.18} & 64.46 & 65.08 \\ \hline
        3& \textcolor{Green}{62.94} & 54.16 & 58.72 & 59.21 \\ \hline
        4& 52.91 & 57.94 & 59.59 & \textcolor{Green}{60.09} \\ \hline
        5& 44.90 & \textcolor{Green}{63.82} & 61.58 & 61.60 \\ \hline
        6& 53.72 & 58.08 & 62.05 & \textcolor{Green}{62.59} \\ \hline
        7& 63.32 & 61.41 & 64.59 & \textcolor{Green}{65.19} \\ \hline
        8& 57.54 & \textcolor{Green}{60.77} & 45.75 & 47.05 \\ \hline
        9& 46.85 & \textcolor{Green}{57.92} & 50.42 & 50.02 \\ \hline
        10& 36.55 & 50.20 & 59.13 & \textcolor{Green}{59.91} \\ \hline
        11& 68.71 & 62.35 & 68.85 & \textcolor{Green}{68.99} \\ \hline
        12& 47.28 & 58.35 & 63.94 & \textcolor{Green}{64.29} \\ \hline
        13& 62.75 & 52.70 & \textcolor{Green}{61.99} & \textcolor{Green}{61.99} \\ \hline
        14& 61.86 & \textcolor{Green}{61.88} & 59.61 & 60.49 \\ \hline
        15& 54.51 & \textcolor{Green}{62.84} & 60.51 & 60.92 \\ \hline
        16& 61.08 & 61.22 & \textcolor{Green}{65.03} & 64.96 \\ \hline
        17& 51.27 & 56.43 & \textcolor{Green}{60.06} & 59.80 \\ \hline
        18& 62.57 & 59.75 & \textcolor{Green}{68.93} & 68.78 \\ \hline
        \hline
        Avg& 56.23 & 59.29 & 61.00 & \textcolor{Green}{61.34} \\ \hline
        STD& 8.43 & 4.13 & 5.63 & 5.49 \\ \hline
    \end{tabularx}
    \caption{`Par. RCSP-Rie'}
\end{minipage}
\end{table*}

\begin{table*}[t]
\centering
\caption{F1 scores (\%) for each testing subject in LOSO-CV classification - Delay Tempo. Highest performing score for each subject is highlighted in green.}
\label{Ind_LOSOCV_Delay}
\begin{minipage}{0.48\linewidth}
    \centering
    \renewcommand{\arraystretch}{1.2}
    \begin{tabularx}{\linewidth}{|p{0.6cm}|X|X|X|X|}
        \hline
        Test Sub & Baseline - w/out PS&LOSO (`Adaptive M') \cite{He_Wu_2020}& LOSO (`TS') \cite{Bleuze_2022}  &LOSO (`ITSA' - \textit{proposed} PS)\\\hline
        1& 59.68 & \textcolor{Green}{61.14} & 53.99 & 53.99 \\ \hline
        2& \textcolor{Green}{65.47} & 62.99 & 58.94 & 58.36 \\ \hline
        3& 47.81 & 55.68 & \textcolor{Green}{57.12} & \textcolor{Green}{57.12} \\ \hline
        4& 45.24 & \textcolor{Green}{52.34} & 49.18 & 49.58 \\ \hline
        5& 54.82 & \textcolor{Green}{67.27} & 60.65 & 60.63 \\ \hline
        6& 22.08 & 45.68 & \textcolor{Green}{65.46} & 65.44 \\ \hline
        7& 53.18 & \textcolor{Green}{56.62} & 53.82 & 53.77 \\ \hline
        8& 16.87 & \textcolor{Green}{59.67} & 48.61 & 48.61 \\ \hline
        9& 23.69 & \textcolor{Green}{61.25} & 46.77 & 46.83 \\ \hline
        10& 8.90 &\textcolor{Green}{52.61} & 51.92 & 51.60 \\ \hline
        11& 53.06 & \textcolor{Green}{59.53} & 59.41 & 59.39 \\ \hline
        12& 60.28 & 57.83 & 62.00 & \textcolor{Green}{63.23} \\ \hline
        13& 34.40 & 49.67 & \textcolor{Green}{56.78} & \textcolor{Green}{56.78} \\ \hline
        14& 64.48 & 63.19 & 64.64 & \textcolor{Green}{64.72} \\ \hline
        15& 17.83 & \textcolor{Green}{60.42} & 58.69 & 58.92 \\ \hline
        16& 37.86 & 47.27 & 56.18 & \textcolor{Green}{56.22} \\ \hline
        17& 56.59 & 56.81 & 61.61 & \textcolor{Green}{61.88} \\ \hline
        18& 33.08 & 44.91 & 62.63 & \textcolor{Green}{63.96} \\ \hline
        \hline
        Avg& 41.96 & 56.38 & 57.13 & \textcolor{Green}{57.28} \\ \hline
        STD& 18.10 & 6.41 & 5.50 & 5.65 \\ \hline
    \end{tabularx}
    \caption{`Seq. RCSP-Rie'}
\end{minipage}
\hfill
\begin{minipage}{0.48\linewidth}
    \centering
    \renewcommand{\arraystretch}{1.2}
    \begin{tabularx}{\linewidth}{|p{0.6cm}|X|X|X|X|}
        \hline
        Test Sub & Baseline - w/out PS&LOSO (`Adaptive M') \cite{He_Wu_2020}& LOSO (`TS') \cite{Bleuze_2022}  &LOSO (`ITSA' - \textit{proposed} PS)\\\hline
        1& 55.97 & \textcolor{Green}{61.63} & 57.06 & 57.46 \\ \hline
        2& 64.49 & \textcolor{Green}{64.50} & 60.56 & 60.80 \\ \hline
        3& 38.84 & 56.41 & 56.82 & \textcolor{Green}{56.96} \\ \hline
        4& 42.46 & \textcolor{Green}{56.29} & 51.63 & 51.97 \\ \hline
        5& 48.66 & \textcolor{Green}{67.58} & 61.56 & 66.56 \\ \hline
        6& 35.56 & 47.53 & \textcolor{Green}{66.15} & 66.09 \\ \hline
        7& 53.62 & 55.23 & 59.27 & \textcolor{Green}{59.30} \\ \hline
        8& 5.45 & \textcolor{Green}{54.94} & 48.50 & 48.51 \\ \hline
        9& 16.79 & \textcolor{Green}{63.02} & 47.57 & 47.55 \\ \hline
        10& 5.39 & 46.63 & \textcolor{Green}{52.89} & 52.55 \\ \hline
        11& 55.53 & \textcolor{Green}{61.51} & 60.31 & 60.17 \\ \hline
        12& 59.87 & 61.73 & 62.24 & \textcolor{Green}{62.35} \\ \hline
        13& 58.11 & \textcolor{Green}{51.87} & 56.52 & 56.52 \\ \hline
        14& 66.38 & 58.44 & 68.65 & \textcolor{Green}{68.69} \\ \hline
        15& 17.56 & 55.01 & 60.33 & \textcolor{Green}{60.54} \\ \hline
        16& 41.02 & 42.25 & 57.58 & \textcolor{Green}{57.92} \\ \hline
        17& 58.80 & 58.24 & 59.80 & \textcolor{Green}{59.85} \\ \hline
        18& 43.10 & 28.74 & \textcolor{Green}{64.59} & 64.56 \\ \hline
        \hline 
        Avg& 42.65 & 55.08 & 58.45 & \textcolor{Green}{58.52} \\\hline
        STD& 19.52 & 9.25 & 5.65 & 5.65 \\ \hline
    \end{tabularx}
    \caption{`Par. RCSP-Rie'}
\end{minipage}
\end{table*}

\subsection{Classifier Sensitivity and Robustness to Hyperparameter Tuning}
To further validate that performance improvements arise from ITSA alignment rather than classifier choice or parameter configuration, we performed hyperparameter tuning of the SVM ($C = 0.01, 0.1, 1, 10$) and additionally evaluated a Random Forest classification on three testing subjects using the best performing feature extraction method (`Par. RCSP-Rie')  in the `Advance Tempo'.  Table \ref{Hyperparamtertune_Table} reports the averaged F1 scores across our LOSO validation using testing subjects 6, 11 and 18. While SVM hyperparameter tuning resulted in marginal improvements, with $C=0.1$ slightly outperforming $C=1$, `ITSA' alignment consistently outperformed unaligned features across all classifiers and parameter configurations.  

\begin{table*}
    \centering
    \begin{tabular}{|c|c|c|c|c|c|}\hline
 & \multicolumn{4}{|c|}{SVM hyperparameter tune}&RF\\ \hline
         &  0.01&  0.1&  1&  10& \\\hline
         Without PS&  56.65 $\pm$ 3.29&  \textcolor{ForestGreen}{61.68 $\pm$ 6.95}&  61.67 $\pm$ 7. 54&  59.80 $\pm$ 9.32 & 49.39 $\pm$
13.72\\\hline
         With ITSA&  61.84 $\pm$ 6.81&  \textcolor{ForestGreen}{66.84 $\pm$ 6.81}&  66.79 $\pm$ 3.64&  63.15 $\pm$ 6.46& 62.97 $\pm$ 6.39\\ \hline
    \end{tabular}
    \caption{Averaged F1 scores for hyperparameter tuned SVM and Random Forest classification on LOSO validation using testing subjects 6, 11 and 18 with `Par. RCSP-Rie' feature extraction in the `Advance Tempo'. F1 scores with  highest performing parameters are denoted in green.}
    \label{Hyperparamtertune_Table}
\end{table*}

\subsection{Additional metrics reported for LOSO-CV}

\begin{figure}[htbp]
    \centering
    \subfloat[]{
        \includegraphics[width=0.45\textwidth]{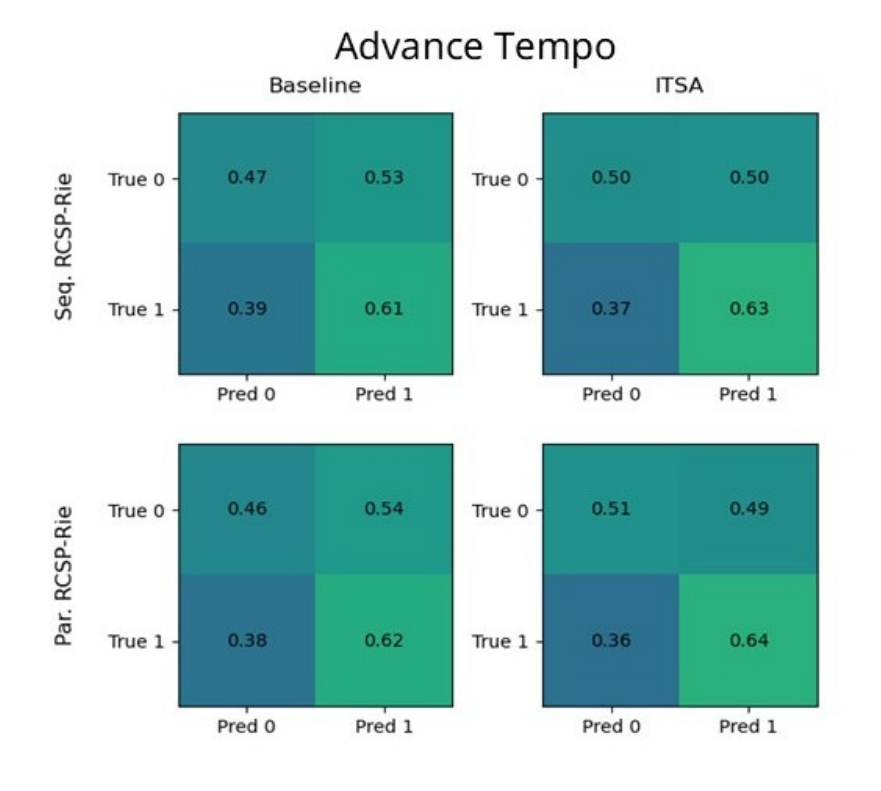}
    }
    \hfill
    \subfloat[]{
        \includegraphics[width=0.45\textwidth]{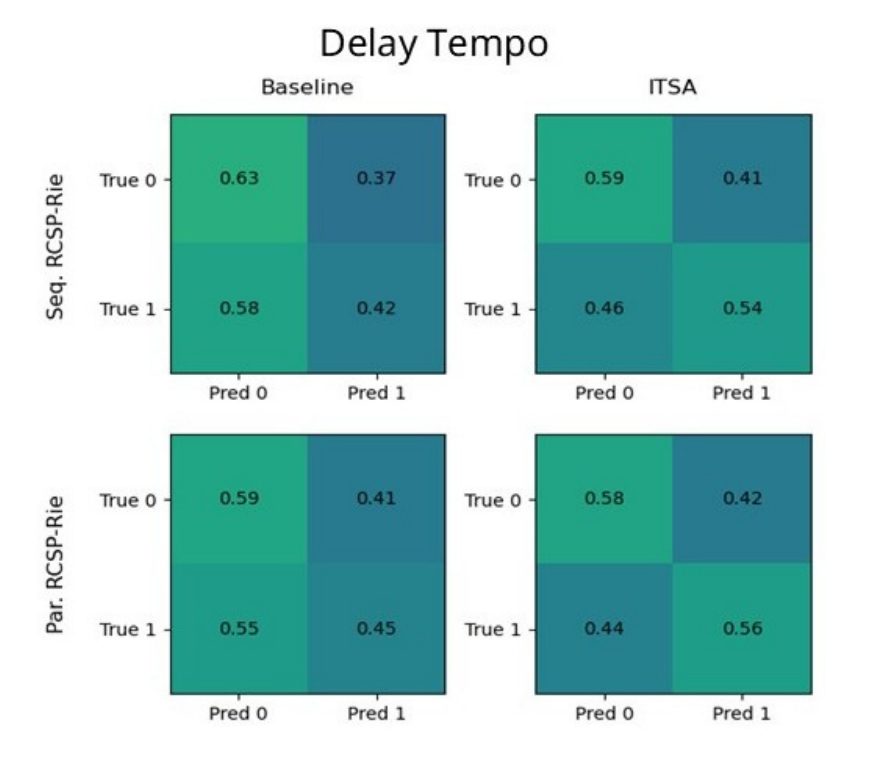}
    }
    \caption{Confusion matrix across temporal conditions `Advance' and `Delay Tempo' and feature extraction methods `Seq.' and `Par. RCSP-Rie' comparing Baseline - unaligned features and proposed `ITSA' implementation.}
    \label{LOSOResults_ConfusionMatrix}
\end{figure}

\begin{figure}[htbp]
    \centering
    \includegraphics[width=0.9\linewidth]{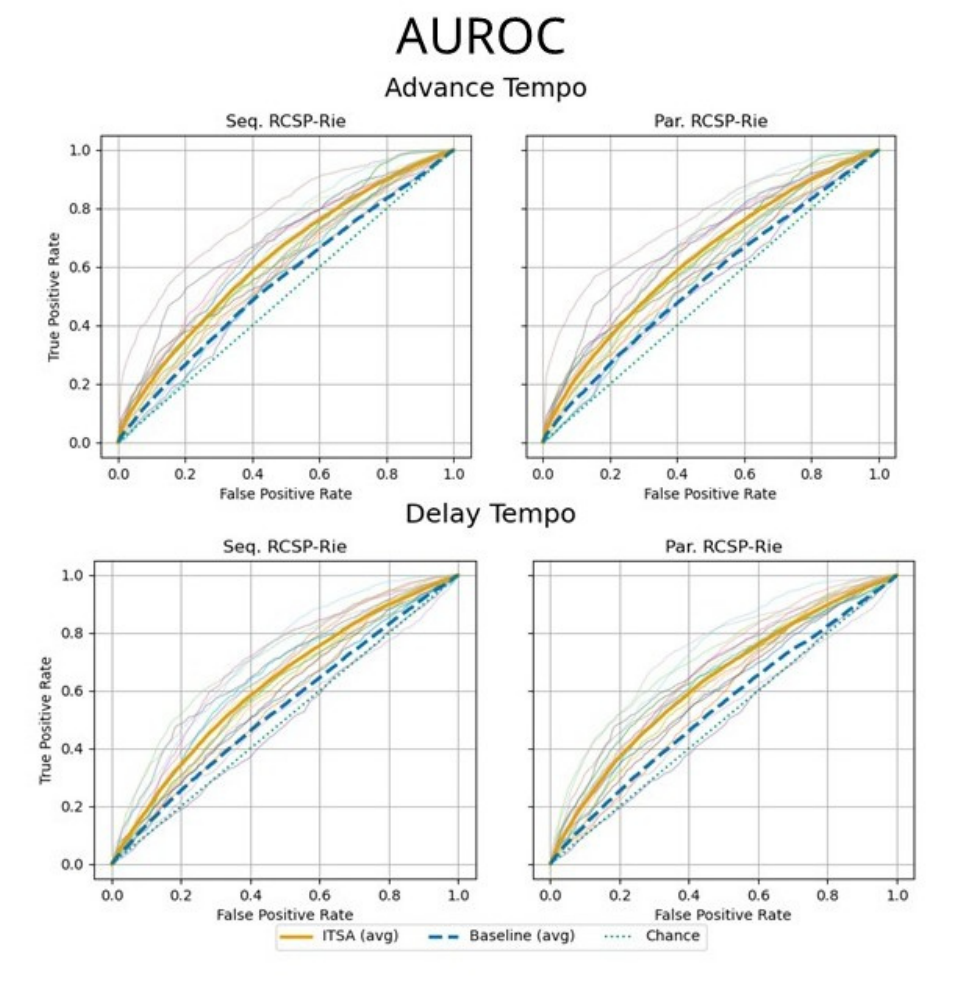}
    \caption{AUROC plots across temporal conditions `Advance' and `Delay Tempo' and feature extraction methods `Seq.' and `Par. RCSP-Rie' comparing \textcolor{blue}{Baseline} - unaligned features (blue, dashed) and proposed \textcolor{orange}{`ITSA'} implementation (orange)} 
    \label{LOSOResults_AUROC}
\end{figure}

\begin{figure}[htbp]
    \centering
    \includegraphics[width=0.9\linewidth]{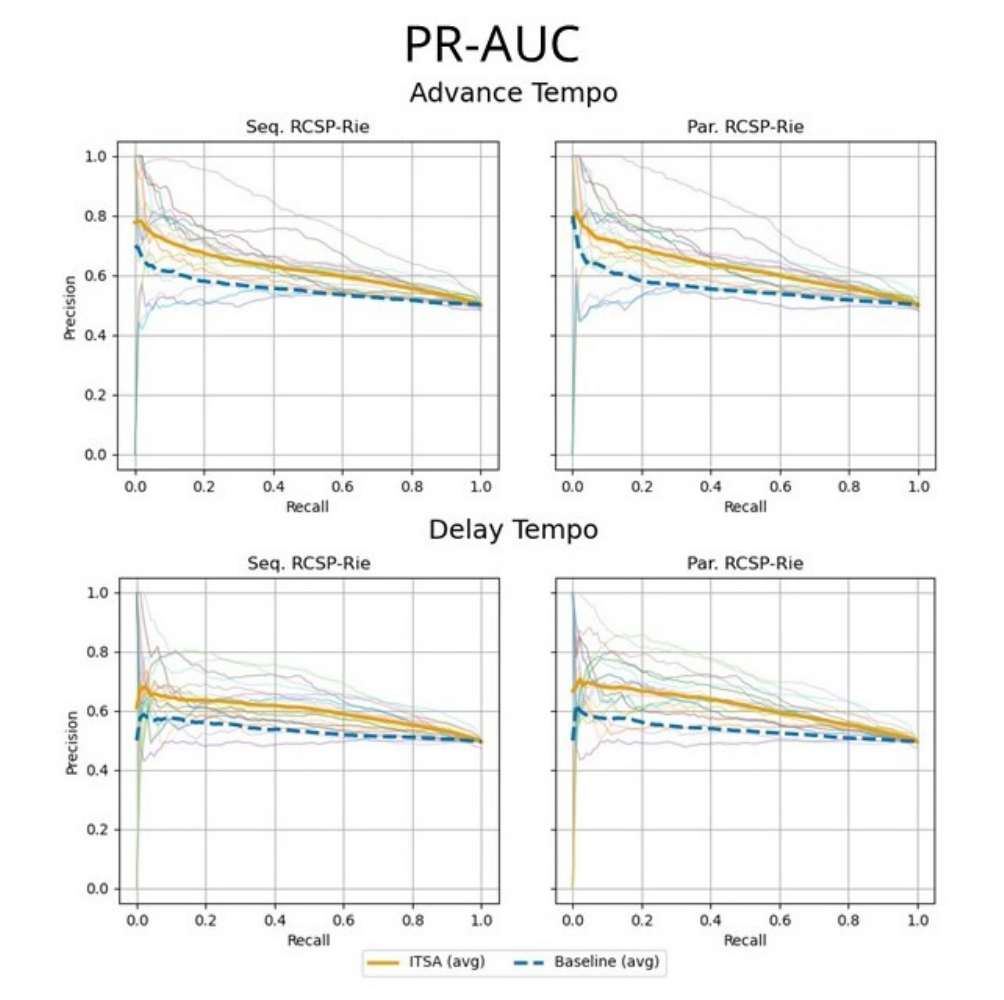}
    \caption{PR-AUC plots across temporal conditions `Advance' and `Delay Tempo' and feature extraction methods `Seq.' and `Par. RCSP-Rie' comparing \textcolor{blue}{Baseline} - unaligned features (blue, dashed) and proposed \textcolor{orange}{`ITSA'} implementation (orange)} 
    \label{LOSOResults_PRAUC}
\end{figure}

Figures \ref{LOSOResults_ConfusionMatrix}, \ref{LOSOResults_AUROC} and \ref{LOSOResults_PRAUC} illustrate the confusion matrices, AUROC and PR-AUC plots for temporal conditions - `Advance' and `Delay' tempo across both feature extraction methods `Seq.' and `Par. RCPS-Rie' comparing performance between Baseline against our proposed `ITSA' implementation.  


Statistical comparisons were conducted across Adaptive M, TS, and ITSA using the Friedman test, followed by pairwise Wilcoxon signed-rank tests with Holm correction for multiple comparisons are reported in Table \ref{mulit_statistical_results}. ITSA achieved the highest mean F1-score across all four experimental conditions. Following Holm correction for multiple comparisons, the statistical analysis provided additional evidence of a statistically significant improvement of ITSA over TS in the Advance Tempo–Par. RCSP-Rie condition ($p=0.0224$).

\begin{table*}[t]
        \centering
        \caption{Statistical comparison of F1-score performance across temporal conditions and feature extraction methods. Values in the post-hoc column are reported as raw $p$-value $\rightarrow$ Holm-adjusted $p$-value. The Friedman $p$-value represents the omnibus test across Adaptive M, TS, and ITSA. Post-hoc comparisons were performed using Wilcoxon signed-rank tests, with the raw $p$-value $\rightarrow$ Holm-adjusted $p$-value, with Holm correction reported across the three pairwise comparisons within each condition. Uncorrected $p<0.05$ are denoted by \textcolor{orange}{(*)} and \textcolor{Goldenrod}{(**)} indicate $p<0.05$ after Holm correction.}
        \label{mulit_statistical_results}
    
        \begin{tabularx}{\textwidth}{
            | >{\centering\arraybackslash}p{2.5cm} |
            >{\centering\arraybackslash}p{2.5cm} |
            >{\centering\arraybackslash}p{2cm} |
            >{\centering\arraybackslash}p{2cm} |
            >{\raggedright\arraybackslash}X |
        }
        \toprule
    
        \textbf{Temporal condition} &
        \textbf{Feature} &
        \textbf{Friedman $\chi^2$(2)} &
        \textbf{Friedman $p$} &
        \textbf{Post-hoc Wilcoxon comparisons} \\
    
        \midrule
    
        \multirow{2}{*}{Advance Tempo}
        & Seq. RCSP-Rie
        & 10.732
        & 0.0047
        &
        \begin{tabular}[t]{@{}l@{}}
            Adaptive M vs.\ TS: 0.0347 \textcolor{orange}{(*)} $\rightarrow$ 0.1040\\
            Adaptive M vs.\ ITSA: 0.0386 \textcolor{orange}{(*)} $\rightarrow$ 0.1040\\
            TS vs.\ ITSA: 0.0552 $\rightarrow$ 0.1040
        \end{tabular}
        \\[4mm]
    
        & Par. RCSP-Rie
        & 5.324
        & 0.0698
        &
        \begin{tabular}[t]{@{}l@{}}
            Adaptive M vs.\ TS: 0.0936 $\rightarrow$ 0.1284\\
            Adaptive M vs.\ ITSA: 0.0642 $\rightarrow$ 0.1284\\
            TS vs.\ ITSA: 0.0075 \textcolor{orange}{(*)}
            $\rightarrow$ 0.0224 \textcolor{Goldenrod}{(**)}
        \end{tabular}
        \\[4mm]
    
        \midrule
    
        \multirow{2}{*}{Delay Tempo}
        & Seq. RCSP-Rie
        & 0.457
        & 0.7957
        &
        \begin{tabular}[t]{@{}l@{}}
            Adaptive M vs.\ TS: 0.9479 $\rightarrow$ 1.0000\\
            Adaptive M vs.\ ITSA: 0.9133 $\rightarrow$ 1.0000\\
            TS vs.\ ITSA: 0.2775 $\rightarrow$ 0.8326
        \end{tabular}
        \\[4mm]
    
        & Par. RCSP-Rie
        & 2.056
        & 0.3577
        &
        \begin{tabular}[t]{@{}l@{}}
            Adaptive M vs.\ TS: 0.3491 $\rightarrow$ 0.6543\\
            Adaptive M vs.\ ITSA: 0.3271 $\rightarrow$ 0.6543\\
            TS vs.\ ITSA: 0.1024 $\rightarrow$ 0.3073
        \end{tabular}
        \\
    
        \bottomrule
        \end{tabularx}
    \end{table*}
\subsection{Effect of calibration subset size during the Rotational Alignment of ITSA} \label{Appendix_CalibrationSize}
Minimising calibration time is critical for practical deployment, therefore we evaluated how reducing the calibration subset size during the Rotational Alignment of ITSA would affect model performance. In results reported in Section \ref{Results}, we partitioned the testing features and assigned 50\% to the calibration subset, leaving 50\% to the evaluation subset. This approximated to 10 and 13 minutes for calibration for the `Advance' and `Delay Tempo', respectively. We explore the effect of reducing the calibration to approximately 5 minutes (allocating 25\% of testing features to calibration subset, resulting in 4.6 and 6.5 minutes calibration in the `Advance' and `Delay Tempo' respectively). Table XII presents the comparison between baseline, unaligned features, ITSA with 25\% of features allocated for calibration and ITSA with 50\% of features assigned for calibration. Reducing the calibration subset size to 25\% results in minor performance loss compared to using 50\% of the feature for calibration in both temporal conditions. However, in the `Advance Tempo', unlike the 50\% condition, no statistically significant improvement over baseline was observed when with the calibration of 5 minutes. In the `Delay Tempo', although the results remained statistically significant (with and without correction for multiple comparisons), using 5 minutes for calibration lead to higher p-values compared to when the calibration time was 10 minutes (5 minutes: p = 0.009, p = 0.002 versus. 10 minutes: p = 0.0021, p = 0.0008 for the `Seq.' and `Par. RCSP-Rie' conditions, respectively). 

\begin{table*}
        \centering
        \caption{Averaged F1-scores and standard deviations across testing subjects for classification of Advance and Delay Tempo features from our LOSO-CV ablation study. We compared calibration subset (allocating 25\% and 50\% of testing features to our calibration subset) during our Rotation step of the proposed `ITSA' against baseline, unaligned features. Metrics are reported for the two feature extraction methods: Seq. RCSP-Rie and Par. RCSP-Rie. 95\% confidence intervals are also provided. Statistically significant improvement from the implementation of PS method  (uncorrected $p<0.05$ are  denoted by \textcolor{orange}{(*)}, corrected $p<0.0167$ using Bonferroni-correction are denoted by \textcolor{Goldenrod}{(**)} ).}
        \label{CalibrationSubset_PS_results}
        \begin{tabularx}{\textwidth}{|l|X|X|X|}\hline
        \multicolumn{4}{|c|}{\textbf{Advance Tempo}}\\\hline
        &Baseline (No PS) & `ITSA' (25\% calibration)  &`ITSA' (50\% calibration) \\ \hline 
        Seq. RCSP-Rie (\%)&54.39  $\pm$ 11.05 (CI: 48.90 - 59.88)& 58.75 $\pm$ 8.40 (CI: 55.27- 62.21) &61.15 $\pm$ 7.27 (CI: 57.54 - 64.77) \textcolor{orange}{(*)}\\ \hline 
        Par. RCSP-Rie (\%)&56.23 $\pm$ 8.43  (CI: 52.03 - 60.42)& 59.57 $\pm$ 5.02 (CI: 57.07 - 62.07) &61.34 $\pm$ 5.49 (CI: 58.60 - 64.07) \textcolor{orange}{(*)} \textcolor{Goldenrod}{(**)} \\ \hline
        \multicolumn{4}{|c|}{\textbf{Delay Tempo}}\\\hline
        &Baseline (No PS) & `ITSA' (25\% calibration)  &`ITSA' (50\% calibration) \\ \hline 
        Seq. RCSP-Rie (\%)& 41.96 $\pm$ 18.10 (CI: 32.96 - 50.97) & 54.59 $\pm$ 5.89 (CI: 51.66 - 57.51)  \textcolor{orange}{(*)}\textcolor{Goldenrod}{(**)}  &57.28 $\pm$ 5.65 (CI: 54.47 - 60.09)  \textcolor{orange}{(*)}\textcolor{Goldenrod}{(**)} 
        \\\hline
         Par. RCSP-Rie (\%)& 42.65 $\pm$ 19.52 (CI: 32.94 - 52.36) & 56.99 $\pm$ 5.88 (CI: 54.07 - 59.92)  \textcolor{orange}{(*)}\textcolor{Goldenrod}{(**)}   & 58.52 $\pm$ 5.65 (CI: 55.71 - 61.33) \textcolor{orange}{(*)} \textcolor{Goldenrod}{(**)} \\\hline
        \end{tabularx}
    \end{table*}

\subsection{Effect of Trial Count on Performance Curves}
\begin{figure}
        \centering
        \includegraphics[width=0.95\linewidth]{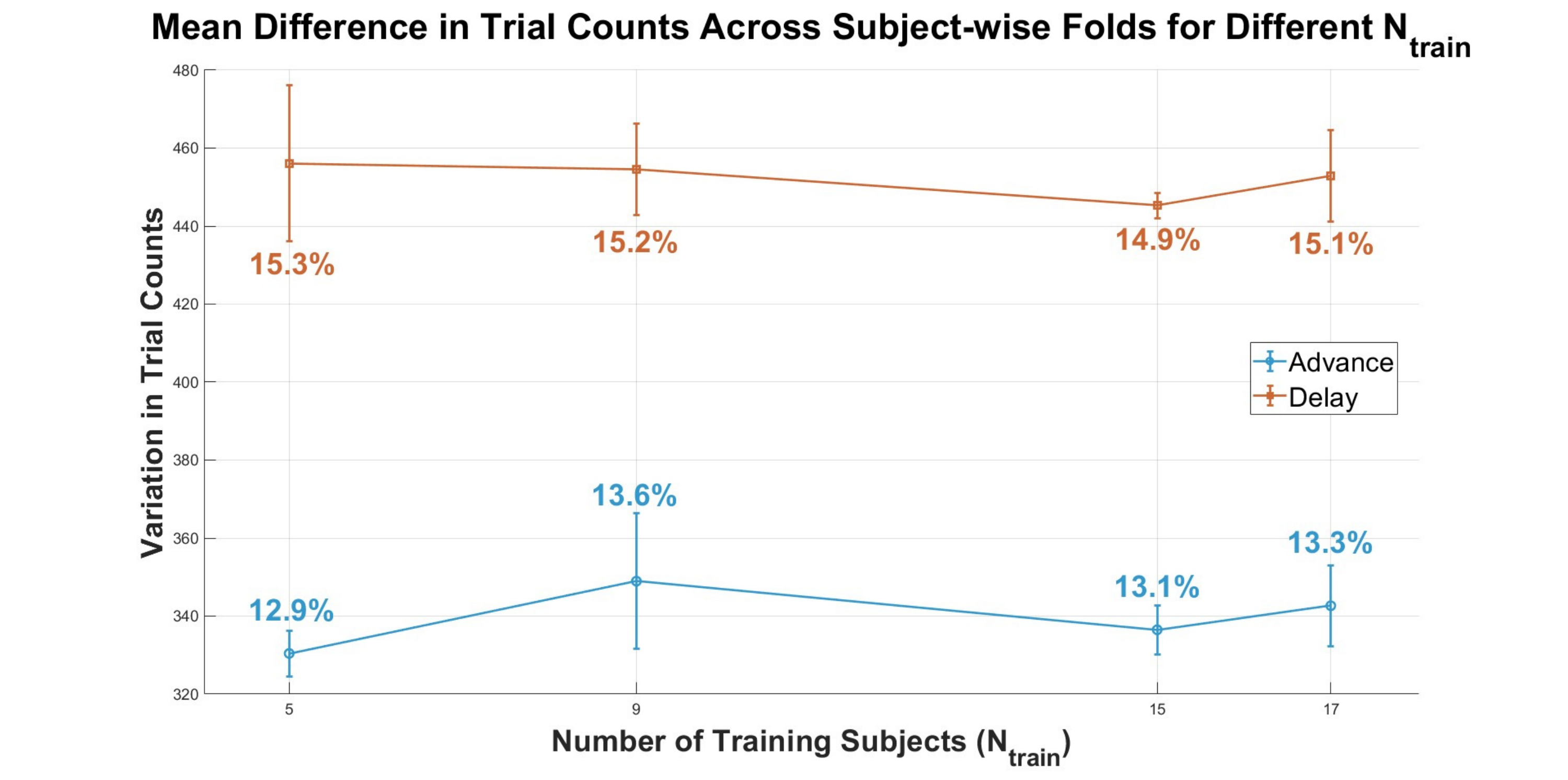}
        \caption{Average variation in trial counts across training subjects for each $N_{train}$  across the temporal conditions `\textcolor{orange}{Advance Tempo}' (orange) and `\textcolor{blue}{Delay Tempo}' (blue). Error bars represent the variability across testing subjects (1, 3, and 18), and percentages for each marker indicate the proportion of overall mean trial count.}
        \label{LearningCurves_VarNumTrials_NTrain_Sub}
    \end{figure}

To examine if the model's performance was driven by the amount of data per subject or by subject diversity, we quantified variation in trial counts across training subjects for different $N_{train}$ as demonstrated in Figure \ref{LearningCurves_VarNumTrials_NTrain_Sub}. Across the testing subjects (and respective training subset folds), the standard deviation of trial counts across training subjects was relatively small (approximately $12-15\%$ of the mean count), indicating that the different in data quantity per subject is unlikely to be the primary driver of performance variability.

\subsection{Feature Reduction } \label{FeatureReductionResults}

\begin{figure}[htbp]
    \centering
    \subfloat[]{
        \includegraphics[width=0.45\textwidth]{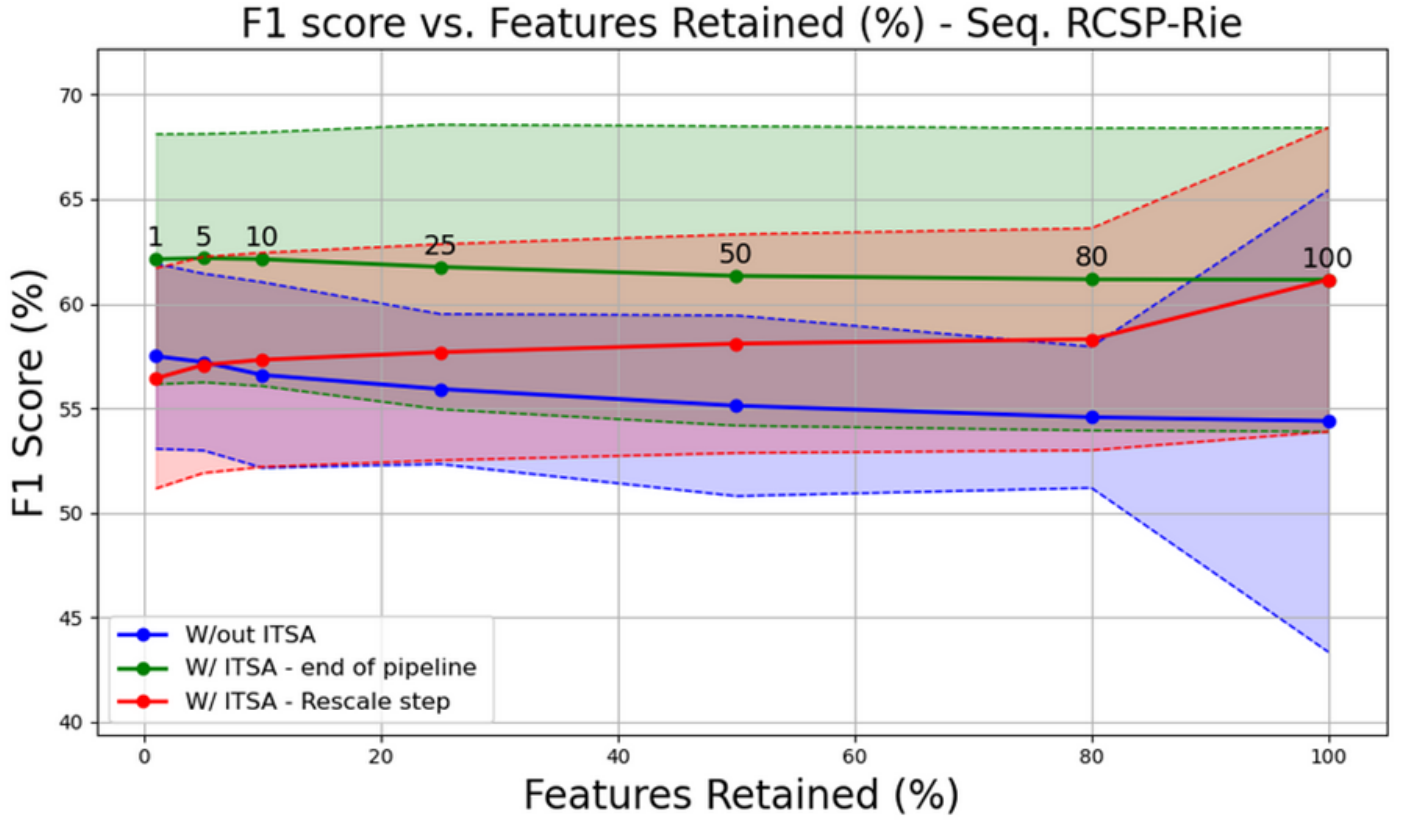}
    }
    \hfill
    \subfloat[]{
        \includegraphics[width=0.45\textwidth]{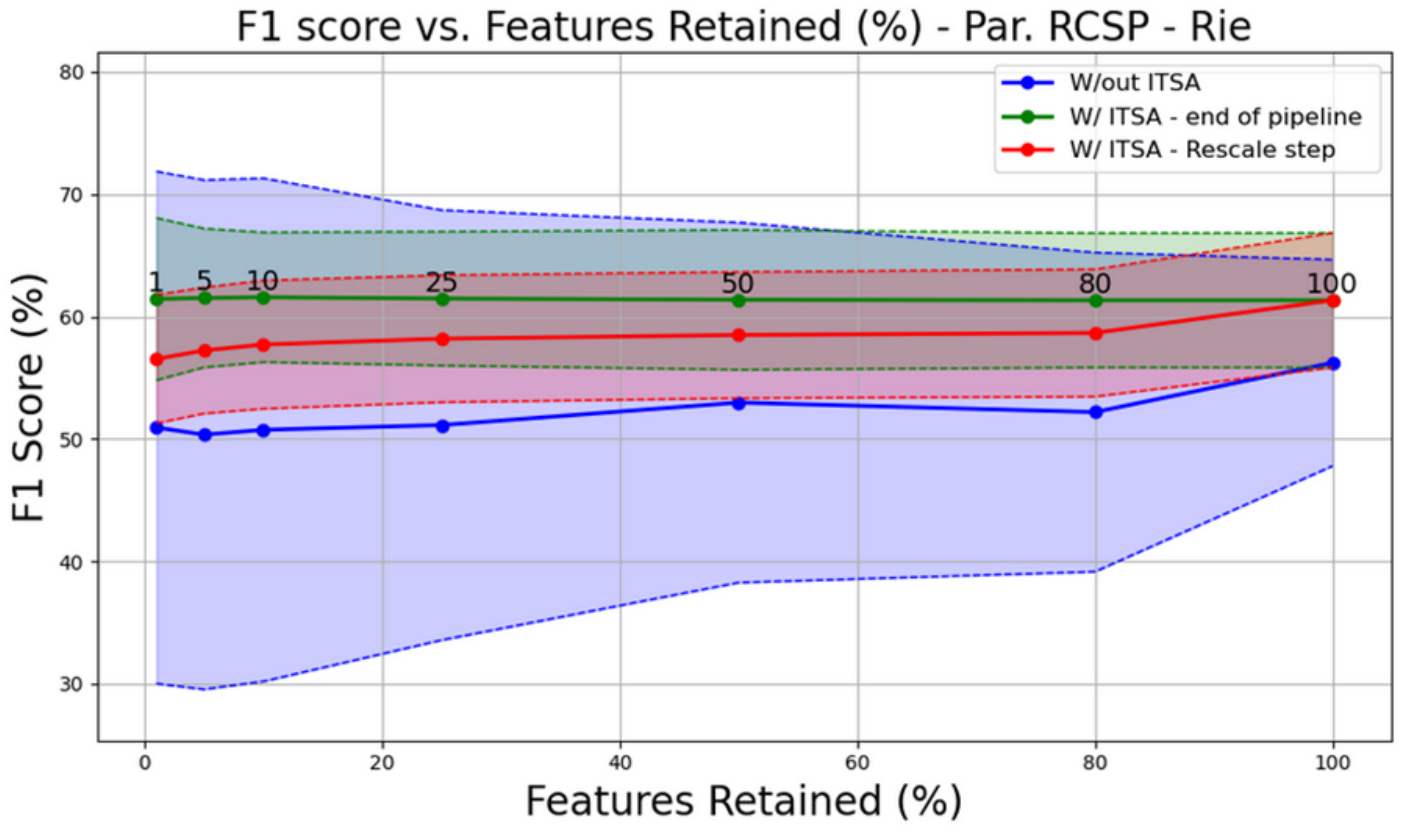}
    }
    \caption{Performance curve illustrating the relationship between the classification performance (F1 score) and the percentage of features retained after application of PCA feature reduction. Averaged results are shown from a LOSO-CV for three configurations: (1) without the use of `ITSA' (`W/out ITSA' - blue), (2) ITSA-aligned features with feature reduction applied at the end of the preprocessing pipeline (`W/ITSA - end of pipeline' - green) and (3) ITSA-aligned features with feature reduction applied during the `rescaling' step (`W/ITSA - Rescale step' - red). Standard deviations are depicted by the shaded area, and respective upper and lower boundaries are displayed with dashed lines.}
    \label{FeatureReduction}
\end{figure}

A feature reduction analysis was conducted to determine the optimal number of features that preserves classification performance while reducing computational demand.
Figure \ref{FeatureReduction} presents the relationship between classification performance (F1 score) and the percentage of features retained after PCA-based dimensionality reduction for both the `Seq.' and `Par RCSP-Rie' approaches in the `Advance Tempo'. The performance curves compare three configurations: (1) baseline features without `ITSA' (`Wout/ ITSA' -  blue), (2) ITSA-aligned features with reduction applied at the end of the pipeline, i.e. after application of RCSP-Rie and ITSA  (`W/ ITSA - end of pipeline'  -  green) and (3) ITSA-aligned features with reduction applied during the `rescaling' step (`W/ITSA - Rescale step' - red). Standard deviations across subjects are represented by the respective shaded areas around the average, with dashed lines illustrating the upper and lower boundaries. 

 The results demonstrate that ITSA-aligned features generally outperformed unaligned features across most reduction levels in both sequential and parallel configurations. Across all three configurations, a positive relationship between classification performance and feature dimensionality is generally observed. 
 An exception is observed with the retention of 1\% of the features in `Seq. RCSP-Rie', where unaligned features marginally outperformed the aligned features that were reduced at the `rescaling' step.  Classification performance for ITSA-aligned features (feature reduced as the rescaling step) remained relatively stable after an initial drop at 80\% feature retention, with the performance loss being modest ($\approx 3\%$). The consistent performance highlights the robustness of the proposed `ITSA' methods, as it suggests that the aligned features would maintain their discriminative power despite the dramatic dimensional reduction to 1\%. 

\subsection{Signal-to-noise comparison across feature extracted from montage configurations}

\begin{figure}
    \centering
    \includegraphics[width=0.95\linewidth]{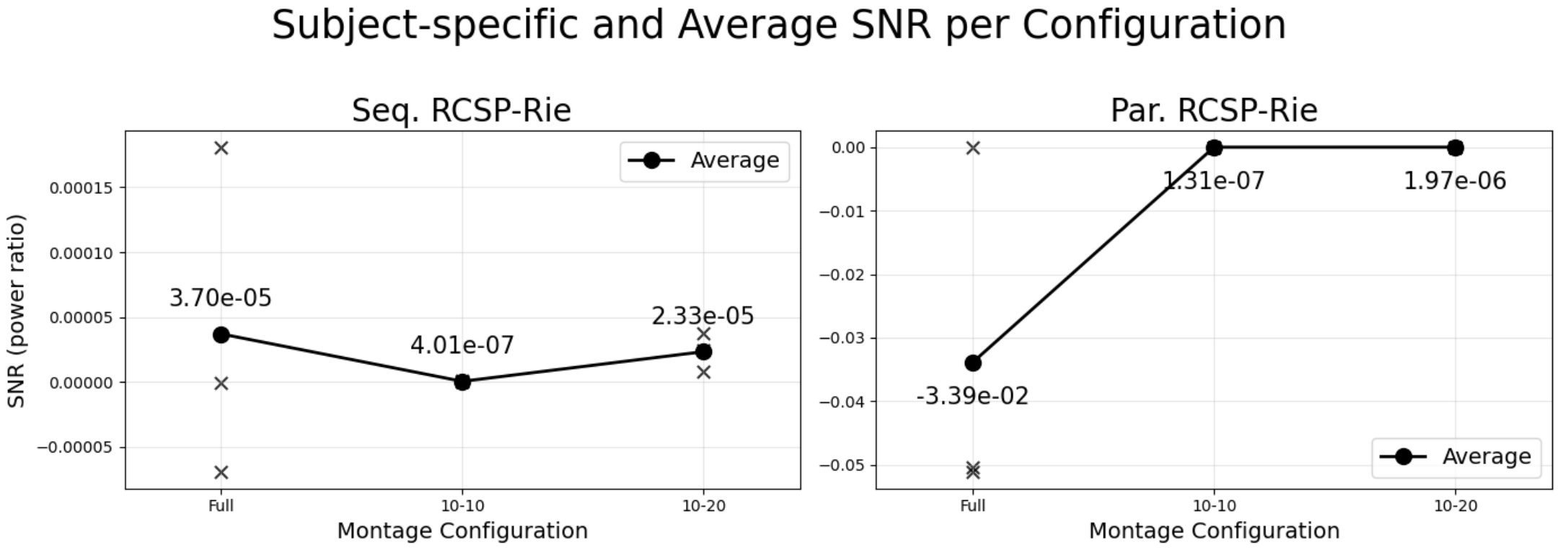}
    \caption{Signal-to-noise ratio (SNR) across testing subjects 1, 6 and 12 for 10-20, 10-10 and full 108-electrode configurations in the Delay Tempo using both `Seq'. and `Par. RCSP-Rie'. Subject-wise and averaged SNR are presented}
    \label{SNR_DelayTempo}
\end{figure}

 For testing subjects 1, 6 and 12, SNR was calculated from the extracted features using `Seq.' or `Par. RCSP-Rie' in the `Delay Tempo'. Our SNR computation consisted of calculating the ratio of the mean to standard deviation across trials for each feature. Specifically for each feature matrix, $X$ of size (trials, $t$ $\times$ features, $f$), we computed: 
\begin{equation}
    SNR = \frac{\text{mean}(X, \text{axis}=0}{\text{std}(X, \text{axis}=0) + \epsilon}
\end{equation}
Where $\epsilon = 10^{-12}$ to prevent division by zero. This approach provided an estimate of signal consistency relative to trial-to-trial variability for each extracted feature. 
 Figure \ref{SNR_DelayTempo} illustrates the subject-wise SNR for each configuration, alongside an averaged SNR across all subjects. Overall, performance comparisons between `Seq.' and `Par. RCSP-Rie' exhibited comparable performance, however the `Par. RCSP-Rie' tended to yield slightly higher performance across temporal conidtions and montage configurations. An increase in SNR was only observed when `Par. RCSP-Rie' from full to 10-10 and 10-20 montages. For the `Seq. RCSP-Rie', we observed a reduction to the variability but no consistent improvement in SNR.  
Additionally, features extracted from the full montage using either `Seq.' or `Par. RCSP-Rie' exhibited greater subject-specific differences compared to the lower configurations, highlighting the influence of individual variability on performance.

\end{document}